\documentclass[runningheads]{llncs}

\usepackage{eccv}

\usepackage{eccvabbrv}

\usepackage{graphicx}
\usepackage{booktabs}

\usepackage[accsupp]{axessibility}  

\usepackage{hyperref}

\usepackage{orcidlink}

\usepackage{ulem}
\newif\ifcolorful

\ifcolorful
  
  \newcommand{\orangetext}[1]{\textcolor{orange}{#1}}

\else
  
  \newcommand{\orangetext}[1]{#1}

\fi

\newcommand{\para}[1]{\vspace{1.5mm}\noindent\textbf{#1}}
\usepackage{graphicx}
\usepackage{subcaption}
\usepackage{multirow}

\begin{document}

\title{ML-SemReg: Boosting Point Cloud Registration with Multi-level Semantic Consistency} 

\titlerunning{ML-SemReg: Boosting PCR with Multi-level Semantic Consistency}

\author{
  Shaocheng Yan\inst{1} \orcidlink{0009-0008-9749-8920} \and
  Pengcheng Shi\inst{2}  \orcidlink{0000-0003-2504-9890} \and
  Jiayuan Li\inst{1, \dag}\orcidlink{0000-0002-9850-1668}
}

\authorrunning{S. Yan et al.}

\institute{
  School of Remote Sensing and Information Engineering, Wuhan University \and
  School of Computer Science, Wuhan University \\
  \email{\{shaochengyan, shipc\_2021, ljy\_whu\_2012\}@whu.edu.cn}
}

\maketitle

\makeatletter
\def\blfootnote{\xdef\@thefnmark{}\@footnotetext}
\makeatother
\blfootnote{$^\dag$ indicates the corresponding author.}

\begin{sloppypar} 

\begin{abstract}
  Recent advances in point cloud registration mostly leverage geometric information. Although these methods have yielded promising results, they still struggle with problems of low overlap, thus limiting their practical usage. In this paper, we propose ML-SemReg, a plug-and-play point cloud registration framework that fully exploits semantic information. Our key insight is that mismatches can be categorized into two types, i.e., inter- and intra-class, after rendering semantic clues, and can be well addressed by utilizing multi-level semantic consistency. We first propose a Group Matching module to address inter-class mismatching, outputting multiple matching groups that inherently satisfy Local Semantic Consistency. For each group, a  Mask Matching module based on Scene Semantic Consistency is then introduced to suppress intra-class mismatching. Benefit from those two modules, ML-SemReg generates correspondences with a high inlier ratio. Extensive experiments demonstrate excellent performance and robustness of ML-SemReg, e.g., in hard-cases of the KITTI dataset, the Registration Recall of MAC increases by almost 34 percentage points when our ML-SemReg is equipped. Code is available at \url{https://github.com/Laka-3DV/ML-SemReg}
\end{abstract}

\section{Introduction}
\label{sec:intro}

\begin{figure}[t]
	\centering
	\includegraphics[width=0.45\linewidth]{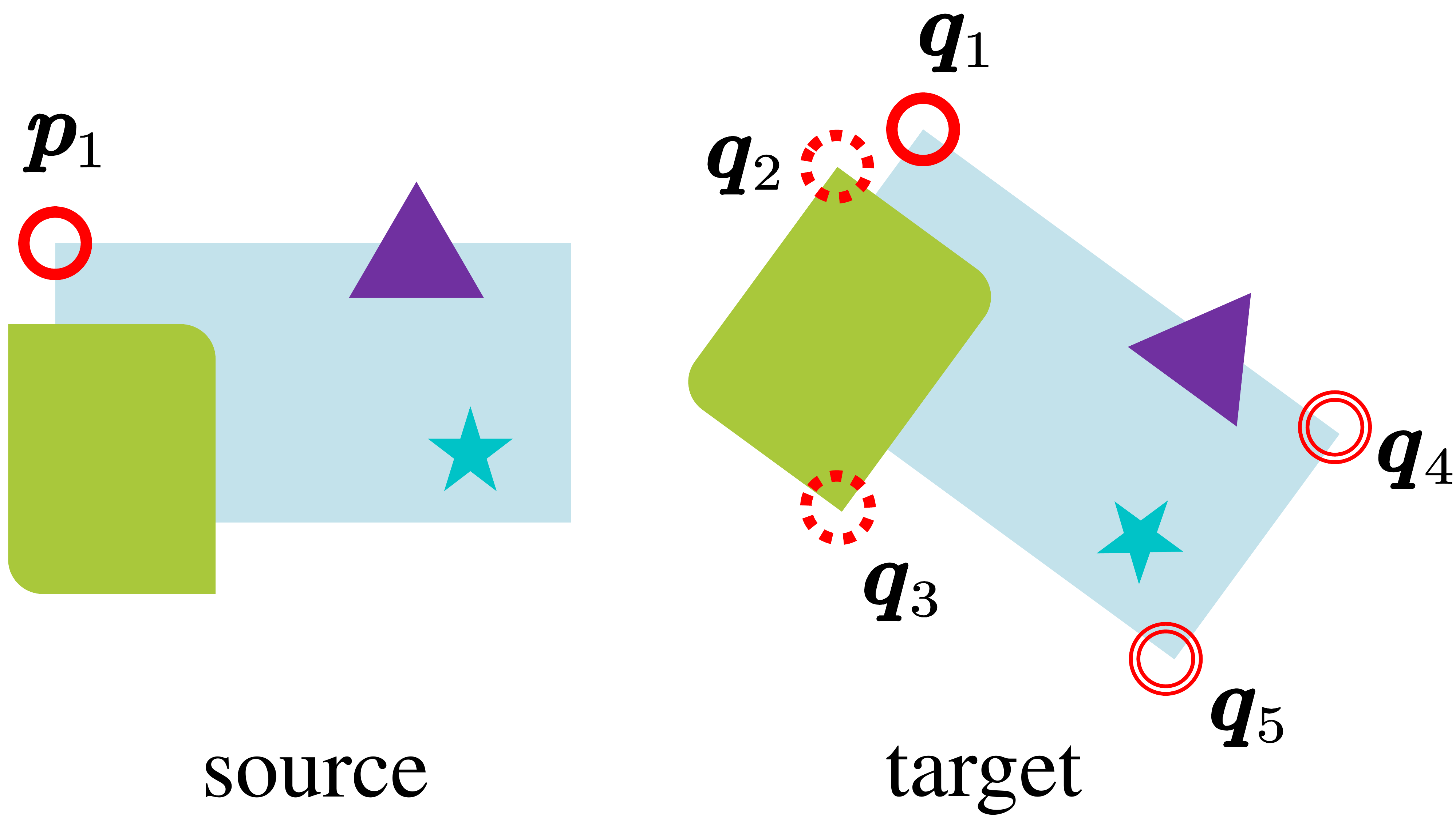}
	\caption{
		Inter-class and intra-class mismatching.
		We use various shapes (such as rectangles, triangles, etc.) to represent different object categories (semantic labels) and depict them in different colors.
		Keypoint (inside the red circle) $\boldsymbol{p}_{1}$ shares the same local geometric features with target keypoints $\boldsymbol{q}_{1\sim 5}$, resulting in {inter-class mismatching} between $\boldsymbol{p}_1$ and $\boldsymbol{q}_{2\sim 3}$, and {intra-class mismatching} between $\boldsymbol{p}_1$ and $\boldsymbol{q}_{4\sim 5}$.
	}
	\label{fig:00}
\end{figure}

Point cloud registration (PCR), a technique that aligns two point clouds from different scans of the same scene, is crucial for applications such as 3D reconstruction \cite{Choi_2015_CVPR}, simultaneous localization and mapping (SLAM) \cite{durrant2006simultaneous}, and augmented reality \cite{azuma1997survey}.
Correspondence-based registration does not rely on initialization and has received more attention recently. It generally consists of two major stages: keypoint matching and transformation estimation\cite{qin2022geometric, yu2023rotation, Poiesi2021, zhang20233d, Chen_2022_CVPR, bai2021pointdsc, wang2022you}.
The latter relies on correspondences outputted by the former. Even though the estimator has achieved good robustness, registration fails when facing small number of inliers or a low inlier ratio\cite{zhang20233d, Chen_2022_CVPR}.
Thus, obtaining high-quality correspondences is an urgent problem to be solved.

Different from image keypoint matching, 3D matching becomes extremely difficult due to problems such as lack of texture, uneven density, and disordered organization \cite{bai2020d3feat, wang2022you}. Therefore, a substantial body of research has concentrated on developing more robust local descriptors, aiming to improve the inlier ratio of correspondences.
\cite{yu2023rotation, Poiesi2021, ao2021spinnet, wang2022you, lepard2021}.
Despite these methods have achieved excellent performance with the development of deep learning, the ubiquitous local geometric similarities in point clouds still pose great challenges to them \cite{Poiesi2021}.
For example, as shown in \cref{fig:00}, keypoint $\boldsymbol{p}_1$ shares identical right-angle local features with candidates $\boldsymbol{q}_{1\sim 5}$, resulting in a correct matching probability of just $\frac{1}{5}$, and the probability of all right angles being perfectly matched is even lower, as low as $\frac{1}{5!} \approx 0.83\%$.

Semantic information can help gain insights into mismatches.
As illustrated in \cref{fig:00}, mismatches after rendering semantic information can be categorized into two types:
(1) Inter-class mismatching: Even though two keypoints are similar in geometric information, they are located in different semantic categories,
e.g., $\boldsymbol{p}_1$ and $\boldsymbol{q}_{2 \sim 3}$.
(2) Intra-class mismatching: Although keypoint pairs are both similar in geometric information and belong to the same semantic category, they are in different scene contexts,
such as $\boldsymbol{p}_1$ and $\boldsymbol{q}_{4 \sim 5}$.
Recently, semantic-assisted 3D matching has gradually received attention \cite{liu2022sarnet, qiao2023g3reg, yin2023segregator,qiao_pagor}.
These methods only match keypoints with the same label, thus reducing inter-class mismatching.
Unfortunately, some inliers may have different semantic labels due to unreliable semantic segmentation.
Therefore, these methods may lose a significant number of inliers, resulting in registration failure.
Furthermore, they cannot address intra-class mismatching as they ignore the scene information of keypoints.

To address the aforementioned challenges, we introduce a PCR framework named ML-SemReg.
Our key insight is that inter- and intra-class mismatches can be effectively addressed by leveraging multi-level semantic consistency.
\orangetext{For local-level,}
we start by introducing a robust Local Semantic Consistency (LS-Consistency), which fully exploits local semantic relations between keypoints.
Next, to tackle inter-class mismatching, we propose a Group Matching (GM) module to categorize keypoints into matching groups that inherently satisfy LS-Consistency.
\orangetext{For scene-level,}
we introduce a descriptor called Binary Multi-Ring Semantic Signature (BMR-SS) to capture scene semantic information. Based on the BMR-SS, a  Mask Matching (MM) module is introduced for each group to maintain Scene Semantic Consistency (SS-Consistency), thereby suppressing intra-class mismatching. Finally, the ML-SemReg framework integrates GM and MM modules, thus ensuring multi-level semantic consistency. In summary, the contributions of our work are three-fold:

\begin{itemize}
    \item A Group Matching module is introduced to address inter-class mismatching, leveraging our robust LS-Consistency.
	\item A Mask Matching module is proposed to address intra-class mismatching based on SS-Consistency by utilizing our scene-aware descriptor BMR-SS.
	\item A plug-and-play PCR framework named ML-SemReg is presented to integrate Group Matching and Mask Matching modules, which ensures multi-level semantic consistency to generate high-quality correspondences.
\end{itemize}

\section{Related Work}

\para{3D Keypoint Matching.}
Different from ICP-type methods  \cite{besl1992method,rusinkiewicz2001efficient,segal2009generalized,bouaziz2013sparse} that establish correspondences by nearest searching in coordinate space, keypoint matching uses descriptor distances in feature space. Traditional approaches  \cite{aiger20084, rusu2009fast, guo2013rops, johnson1999using,frome2004recognizing,tombari2010unique,guo2015novel, chen20073d,rusu2008aligning, zaharescu2009surface} rely on manually designed local patch geometric descriptors. With the advancement of deep learning, learned 3D feature descriptors have outperformed hand-crafted ones. A pioneering work is 3DMatch  \cite{zeng20173dmatch}, which uses a siamese convolutional network for description. Recently, many networks have attempted to improve the performance by resorting to rotation invariant modules  \cite{deng2018ppfnet, yang2018foldingnet, deng2018ppf, ao2021spinnet, wang2022you}, fully convolution modules  \cite{choy2019fully, Poiesi2021, lepard2021}, and transformer modules  \cite{yu2023rotation, qin2022geometric, yew2022regtr}, or by designing detect-description  \cite{bai2020d3feat} and encoding-decoding frameworks  \cite{yu2021cofinet,qin2022geometric, huang2021predator}. Although these methods have achieved remarkable performance improvements, they still face numerous challenges caused by local geometric similarities.

\para{Robust Model Estimation.}
Traditional robust estimation methods, such as RANSAC  \cite{fischler1981random}, tend to converge slowly and exhibit instability when dealing with a high proportion of outliers. To address this issue, several algorithms  \cite{yin2023segregator, Yang20tro-teaser, zhang20233d, yang2023mutual} employ length consistency to construct graphs for searching for a maximal clique of inliers and subsequently use robust estimators for transformation estimation. Deep estimators  \cite{choy2020deep, pais20203dregnet, bai2021pointdsc} typically consist of an inlier/outlier classification network and a transformation estimation network, which have shown improvements in both accuracy and speed. While these methods excel at handling low inlier ratios, they may encounter challenges when the number of initial correspondences is extremely small. In such cases, there are too many outliers to vote for inliers \cite{yang2023mutual, zhang20233d}, or the maximum clique assumption might not hold \cite{qiao2023g3reg}, rendering these algorithms ineffective. In contrast, our approach effectively retains inliers and achieves a high inlier ratio, thereby enhancing the robustness and performance of these methods.  

\para{Semantic-assisted Matching.}
Existing solutions can be categorized into keypoint-level and entity-level methods. Keypoint-level methods directly utilize the semantic labels of keypoints to avoid matching between different categories \cite{arce2023padloc, liu2022sarnet, chen2019iros, DBLP:journals/ral/ZaganidisSDC18, DBLP:conf/dicta/TruongGIS19}. For instance, SARNet \cite{liu2022sarnet} employs a semantic mask to filter out correspondences with different categories. In contrast, entity-level algorithms first construct semantic clusters (or entities) and then perform matching on them \cite{qiao_pagor, qiao2023g3reg, yin2023segregator, liu2024deep}. They fully exploit the saliency of entities in the scene, thus being more robust to semantic noise. Segregator \cite{arce2023padloc} combines both entity-level and keypoint-level approaches, incorporating the advantages of both.  However, these methods ignore the scene information of keypoints, making them very sensitive to intra-class mismatching. In contrast, our method explores Scene Semantic Consistency, thereby effectively addressing intra-class mismatching problem. 
\section{Method}
\label{sec:method}

\begin{figure}[tb]
	\centering
	\includegraphics[width=1.0\linewidth]{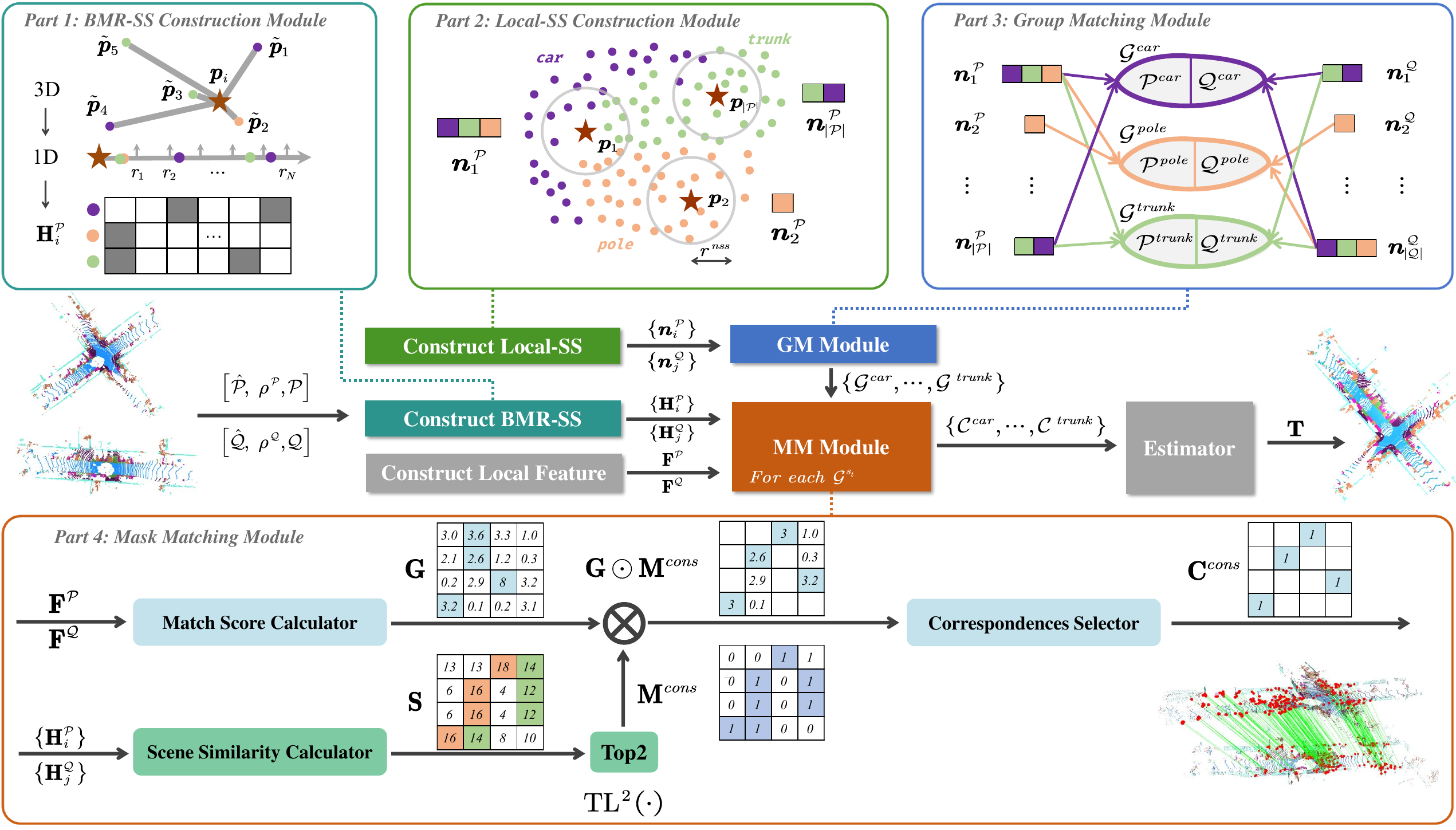}
	\caption{
		{ML-SemReg Framework}. 
		For keypoints from the source and target point clouds, the proposed ML-SemReg 
		(1) constructs the BMR-SS and Local-SS of each keypoint to perceive scene and local semantic information, respectively.
		(2) Simultaneously, the local geometry feature is calculated by a descriptor (e.g., FPFH \cite{rusu2009fast}). 
		(3) For each semantic category, the GM module produces a matching group, in which the keypoints satisfy LS-Consistency among each other.
		(4) For each matching group, the MM module constructs a scene consistency mask, outputting sub-correspondences sets with multi-level semantic consistency. (4) Finally, the union of subsets produces high-quality correspondences, which are used to estimate the rigid transformation matrix.
	}
	\label{fig:SemanReg_arch}
\end{figure}

\subsection{Problem Formulation}
Given two point clouds $\hat{\mathcal{P}} \subset \mathbb{R}^3$ (source) and $\hat{\mathcal{Q}} \subset \mathbb{R}^3$ (target), our goal is to find a sufficient number of inliers to align them. The framework of the ML-SemReg is shown in \cref{fig:SemanReg_arch}, which takes $\hat{\mathcal{P}}$, $\hat{\mathcal{Q}}$, $\rho ^{{\mathcal{P}}}$, $\rho^{{\mathcal{Q}}}$, $\mathcal{P}$, and $\mathcal{Q}$ as its inputs, where $\mathcal{P} = \{\boldsymbol{p}_i\}$ and $\mathcal{Q} = \{\boldsymbol{q}_j \}$ are  keypoints randomly selected from $\hat{\mathcal{P}}$ and $\hat{\mathcal{Q}}$, respectively. $\rho ^{\mathcal{T}}:\mathcal{T}\rightarrow \mathcal{S}, \mathcal{T}\in \{ {\mathcal{P}},{\mathcal{Q}} \}$ represents label mapping function (e.g., PTv2 \cite{wu2022point}), where $\mathcal{S}=\{ s_t \}$ is a set of semantic labels. Our method comprises the Group Matching module (\cref{subsec:group_matching}) and the Mask Matching module (\cref{subsec:mask_matching}), which address inter-class and intra-class mismatching, respectively.
\subsection{Group Matching}
\label{subsec:group_matching}
To solve inter-class mismatching, previous works \cite{arce2023padloc, liu2022sarnet, chen2019iros, he2021vi} strictly avoid keypoint matching between two categories.  However, this approach is not robust when faced with semantic noise, leading to a poor number of inliers and ratio. We address this problem by relaxing the matching of keypoints \textit{within the same category} to a more robust \textit{within matching groups}.  

\para{Local Semantic Consistency.}
We first propose LS-Consistency as follows:
\begin{equation}
    \boldsymbol{n}_{i}^{\mathcal{P}}\cap \boldsymbol{n}_{j}^{\mathcal{Q}}\ne \varnothing \rightarrow \varGamma ^{LS}\left( \boldsymbol{p}_i,\boldsymbol{q}_j \right) =1.
	\label{eq:fafdaaq}
\end{equation}
where $\boldsymbol{n}_{i}^{\mathcal{P}}=\left\{ \rho ^{\mathcal{P}}\left( \boldsymbol{p}_j \right) |\lVert \boldsymbol{p}_j-\boldsymbol{p}_i \rVert_2 \le r^{local} ,\boldsymbol{p}_j\in \mathcal{P} \right\}$ denotes the Local Semantic Signature (Local-SS) under search radius $r^{local}$. The indicator function $\varGamma^{LS} : \mathcal{P} \times \mathcal{Q} \rightarrow \{0, 1\}$ signifies whether points $\boldsymbol{p}_i$ and $\boldsymbol{q}_j$ satisfy the LS-Consistency.  

\para{Group Matching.}
It is easy to prove that when the search radius $r^{local}$ is large enough (e.g., 2 $\times$ the inlier threshold), a correspondence can be classified as an inter-class mismatching (outlier) if two keypoints cannot satisfy the LS-Consistency.  Thus, we can solve inter-class mismatching by avoiding the matching of keypoints that do not satisfy LS-Consistency. To implement this, we first define a Matching Group as a combination of two keypoint sets:
\begin{equation}
	\mathcal{G}^{s_t}= (\mathcal{P}^{s_t},\mathcal{Q}^{s_t}), 
\end{equation}
where $\mathcal{P}^{s_t}=\left\{ \boldsymbol{p}_i|s_t\in \boldsymbol{n}_{i}^{\mathcal{P}},\ \boldsymbol{p}_i\in \mathcal{P} \right\}$ is a sub keypoint set of source keypoints of label $s_t$, and $\mathcal{Q}^{s_t}$ is a sub keypoint set of target keypoints. Obviously, keypoints in $\mathcal{G}^{s_t}$ satisfy LS-Consistency because each of their Local-SS contains $s_t$. Conversely, if two keypoints do not exist in any matching group, they cannot be matched because they do not satisfy LS-Consistency. Therefore, by just matching keypoints in each Matching Group and then merging them, a final set of correspondences without inter-class mismatching can be obtained:
\begin{equation}
	\mathcal{C}=\bigcup_{\begin{array}{c}
		s_t\in \mathcal{S}\\
	\end{array}}{\textrm{MF}\left( \mathcal{G}^{s_t} \right)},
	\label{eq:ss_same}
\end{equation}
where $\text{MF}(\cdot)$ represents a matching function like the nearest neighbor matcher.
\subsection{ Mask Matching}
\label{subsec:mask_matching}

In this section, we first construct the Binary Multi-ring Semantic Signature (BMR-SS) to quantify scene similarity.
Based on scene similarity, we propose the Mask Matching (MM) module to mitigate intra-class mismatching by selectively discarding correspondences involving keypoint pairs with low scene similarity.

\para{Construction of BMR-SS}.
Inspired by \cite{sift2012, ao2021spinnet}, which construct descriptors through spatial division, we adopt this strategy to construct BMR-SS for encoding scene information.
We first construct $Z^{\mathcal{P}}$ clusters of semantic instances (car, traffic sign, pole, etc.) using euclidean clustering with different radii\cite{yin2023segregator}. 
Then, the landmark set is defined as the collection of cluster centers:
\begin{equation}
	\mathcal{L}^{\mathcal{P}}=\{ \tilde{\boldsymbol{p}}_i |i=1\cdots Z^{\mathcal{P}}, \tilde{\boldsymbol{p}}_i \in \mathbb{R}^3 \},
\end{equation}
where $\tilde{\boldsymbol{p}}_i$ is centroid of $i$-th cluster. 
Next, we divide the scene into $N$ rings of width $L$ centered on keypoint,
where the radius range of the $i$-th ring is $[r_{i-1}, r_i], r_i=iL,\forall i=1\cdots N$.
Then, the BMR-SS $\mathbf{H}_i^{\mathcal{P}}\in \{0, 1\}^{ \vert \mathcal{S} \vert \times N }$ is filled by:
\begin{equation}
	[\mathbf{H}_i^{\mathcal{P}}]_{tk} =\begin{cases}
		1 & \text{if } {s_t}\in \varUpsilon _{k}\left( \boldsymbol{p}_i \right) \\
		0 & \text{otherwise}                                                                         \\
	\end{cases}, \quad t=1,\cdots, |\mathcal{S}|, \quad k=1,\cdots, N, 
    \label{eq:fdjhh}
\end{equation}
where 
$
\varUpsilon _k\left( \boldsymbol{p}_i \right) =\{ \rho ^{\mathcal{P}}\left( \tilde{\boldsymbol{p}} \right) |r_{k-1} \le \lVert \tilde{\boldsymbol{p}}-\boldsymbol{p}_i \rVert _2 < r_{k},\tilde{\boldsymbol{p}}\in \mathcal{L}^{\mathcal{P}} \}
$ query the label set of landmark within the $k$-th ring around $\boldsymbol{p}_{i}$.
The interpretation of \cref{eq:fdjhh} is that if a landmark with category $s_{t}$ is in the $k$-th ring of $\boldsymbol{p}_{i}$, then $[\mathbf{H}_{i}^{\mathcal{P}}]_{tk}=1$.
Note that the encoding $\mathbf{H}_{i}^{\mathcal{P}}$ relies solely on distance (1D), thus inherently possessing rotation invariance required for descriptors \cite{Poiesi2021, zeng20173dmatch, deng2018ppf}.

\begin{figure}[t!]
	\centering
	\includegraphics[width=0.999999\linewidth]{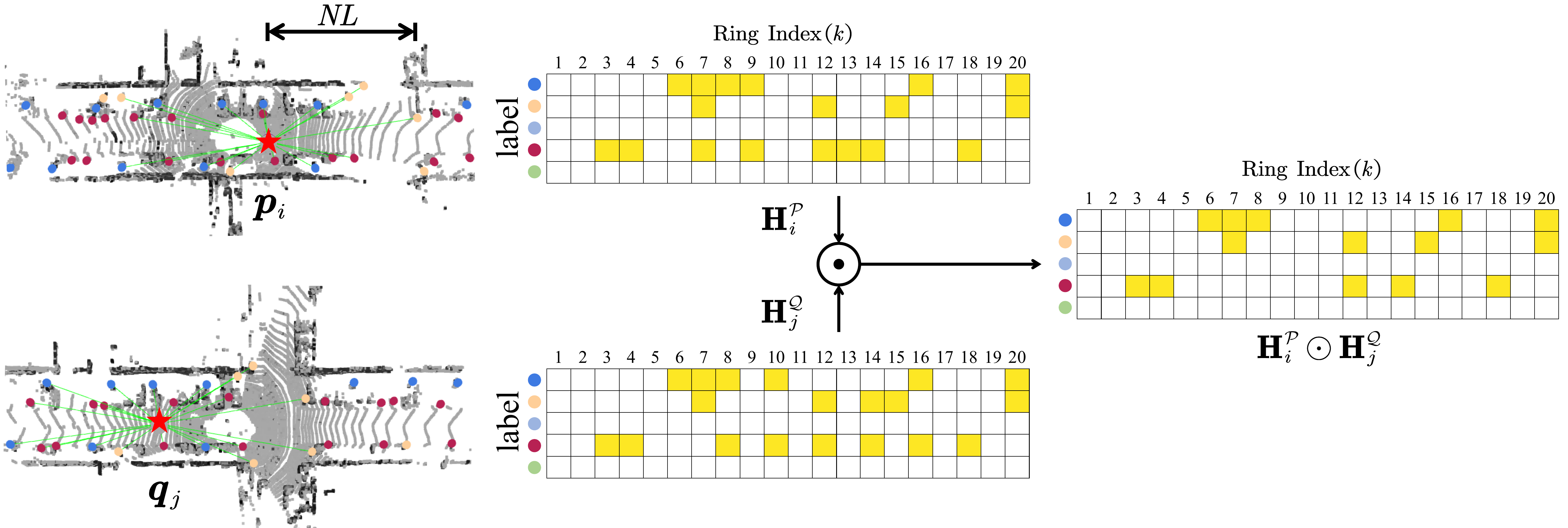}
	\caption{
	Same locations exhibit scene similarity across different LiDAR scans.
	Keypoint $\boldsymbol{p}_i$ is connected to the landmarks (centroid of object) within its maximum receptive field radius $NL$.
	For $\mathbf{H}_i^{\mathcal{P}}$, the $t$-th row and $k$-column correspond to semantic label $s_t$ and $k$-th ring, respectively.
	The final scene similarity between $\boldsymbol{p}_i$ and $\boldsymbol{q}_j$ is $14$ as $\varTheta \left( \boldsymbol{p}_i,\boldsymbol{q}_j \right) =\left| \mathbf{H}_{i}^{\mathcal{P}}\odot \mathbf{H}_{j}^{\mathcal{Q}} \right|=14$.
	}
	\label{fig:03}
\end{figure}

\para{Scene Semantic Similarity.}
Semantic instances in a scene can provide clues for scene similarity. For instance, if two keypoints are both approximately 2 meters away from a traffic light, they may be in similar positions. Building upon this observation, we propose to measure the scene semantic similarity of two keypoints by counting occurrences of the same landmark label within the same distance range around them.
Moreover, if there exists a semantic label $s_t$ in the $k$-th ring for both $\boldsymbol{p}_i$ and $\boldsymbol{q}_j$, then $[\mathbf{H}_{i}^{\mathcal{P}}]_{tk}=[\mathbf{H}_{j}^{\mathcal{Q}}]_{tk}=1$. 
Therefore, the scene semantic similarity between $\boldsymbol{p}_i$ and $\boldsymbol{q}_j$ can be expressed as:

\begin{equation}
	\varTheta \left( \boldsymbol{p}_i,\boldsymbol{q}_j \right) =\sum_{t=1}^{\left| \mathcal{S} \right|}{\sum_{k=1}^{N}{\left[ \mathbf{H}_{i}^{\mathcal{P}} \right] _{tk}\times \left[ \mathbf{H}_{j}^{\mathcal{Q}} \right] _{tk}}}=|\mathbf{H}_{i}^{\mathcal{P}}\odot \mathbf{H}_{j}^{\mathcal{Q}}|.
\end{equation}
\cref{fig:03} illustrates the calculation of scene semantic similarity between $\boldsymbol{p}_i$ and $\boldsymbol{q}_j$ (the same location in different LiDAR scans) using BMR-SS.

\para{Mask Matching.}
The MM module aims to integrate scene semantic similarity into local feature matching, addressing intra-class mismatching.
To combine the two levels of features, we initially formalize feature matching methods as a selection process, utilizing a match score matrix as input:
\begin{equation}
	\mathbf{C} = \theta(\mathbf{G}), 
	\label{eq:Dj94oaa}
\end{equation}
\orangetext{where $\mathbf{G} \in \left[ 0,1 \right] ^{ \vert \mathcal{P} \vert \times \vert \mathcal{Q} \vert }$ denotes the match score matrix based on local descriptor similarities, such as the cosine similarity of the GEDI descriptor.}
$\theta: \left[ 0,1 \right] ^{ \vert \mathcal{P} \vert \times \vert \mathcal{Q} \vert } \rightarrow \{0,1\}^{ \vert \mathcal{P} \vert \times \vert \mathcal{Q} \vert}$ is 
binary function, which selects matches with high match scores, outputting $\mathbf{C} \in \{0,1\}^{ \vert \mathcal{P} \vert \times \vert \mathcal{Q} \vert }$, where $\mathbf{C}_{ij}=1$ means $(\boldsymbol{p}_i, \boldsymbol{q}_j)$ is a putative inlier.
Subsequently, by reweighting the match score matrix using an elaborate scene consistency mask $\mathbf{M}^{cons}\in\{0,1\}^{|\mathcal{P}| \times |\mathcal{Q}|}$, we can effectively suppress intra-class mismatching.
To construct $\mathbf{M}^{cons}$,
We first calculate the scene similarity matrix $\mathbf{S} \in \mathbb{N}^{\left|\mathcal{P}\right| \times \left|\mathcal{Q}\right|}$,
where $ [\mathbf{S}]_{ij} = \varTheta(\boldsymbol{p}_i, \boldsymbol{q}_j)$ represents the scene similarity between keypoints $\boldsymbol{p}_i$ and $\boldsymbol{q}_j$.
Next, a binary selection function
$\text{TL}^K(\cdot)$ is used to set the highest $K$ values in each row of $\mathbf{S}$ to $1$, while the remaining values are set to $0$.
This process yields the scene consistency mask $\mathbf{M}^{cons}$, where $[\mathbf{M}^{cons}]_{ij} = 1$ means that $\boldsymbol{p}_i$ and
$\boldsymbol{q}_j$ satisfy \orangetext{Scene Semantic Consistency (SS-Consistency)}.
Finally, based on the masked match score matrix $\mathbf{G}\odot\mathbf{M}^{cons}$, correspondence selection can be represented as an extension of~\cref{eq:Dj94oaa}:
\begin{equation}
	\mathbf{C}^{cons}=\theta \left( \mathbf{G}\odot \mathbf{M}^{cons} \right).
\end{equation}
By applying Mask Matching to each group generated by the Group Matching module, we can produce a high-quality set of correspondences with 
both local and Scene Semantic Consistency:

\begin{equation}
	\mathcal{C}=\bigcup_{\begin{array}{c}
		\scriptstyle \mathcal{G}^{s_t}\in \textrm{GM}\left( \mathcal{P},\mathcal{Q}\right) \\
	\end{array}}{\textrm{MM}\left( \mathcal{G}^{s_t} \right)}.
\end{equation}
\section{Experiment}
\label{sec:experiment}

\subsection{Datasets and Experimental Setup}

\para{Outdoor Benchmark.}
The KITTI-Odometry dataset \cite{Geiger2012CVPR} is used for outdoor evaluation.
Following  \cite{bai2021pointdsc, zhang20233d, choy2019fully}, we construct an easy-type dataset using point cloud pairs with a $10$m translation offset.
Additionally, we construct medium-type and hard-type datasets with translational distances of $20$m and $30$m, respectively, which can comprehensively assess registration performance in scenarios with low overlap ratios.
Note that we conduct the evaluation on all KITTI sequences instead of only sequences 8$\sim$10  \cite{choy2019fully, zhang20233d, Chen_2022_CVPR, qin2022geometric}.
Hence, we obtain $2096$, $1060$, and $702$ point cloud pairs for registration in these three types of datasets.

\para{Indoor Benchmark.}
The ScanNet dataset  \cite{dai2017scannet} is used for indoor evaluation. 
Since the raw ScanNet dataset is huge, we apply a selection criterion to retain pairs whose overlap ratios falling within a specific range in the sub-dataset \textit{scannet\_frames\_25k}, resulting in $1142$ point cloud pairs with overlap ratios ranging from $50\%$ to $60\%$ and $980$ pairs with overlap ratios ranging from $40\%$ to $50\%$.

\para{Implementation Details.}
Our algorithm is implemented in Python using the Open3D library  \cite{Zhou2018open3d}. We use PTv2 \cite{wu2022point} to generate semantic labels. In the Group Matching module, $r^{local}$ is set to $0.8m$ for KITTI and $5cm$ for ScanNet. During the construction of BMR-SS, the number of ring ($N$) and the length of ring ($L$) are set to $33$ and $1.5m$ for KITTI, while $10$ and $20cm$ for ScanNet. In the MM module, $K$ is set to 2 and 3 for KITTI and ScanNet, respectively. All experiments are conducted on a PC equipped with an Intel(R) i7-13700KF processor, 32GB of RAM, and a GeForce RTX 4090 GPU.  

\para{Evaluation Criteria.}
Following  \cite{choy2019fully},
we evaluate the performance of correspondence set using {Inlier Ratio} (IR) \cite{qin2022geometric, Poiesi2021}.
We observed that removing potential outliers can improve IR. However, this process may also inadvertently reduce the number of inliers, which can negatively affect registration\cite{zhang20233d}. Therefore, we also report the Inlier Number (IN).
For registration evaluation, following  \cite{zhang20233d, Chen_2022_CVPR},
we employ mean Rotation Error (RE), and Translation Error (TE) of successful registered cloud pairs as the evaluation metrics.
We use Registration Recall (RR) to denote the registration success rate, which is defined as the fraction of point cloud pairs with RE$\le 5^\circ$, TE $\le 60cm$ on the KITTI, and RE$\le 15^\circ$, TE $\le 30cm$ on the ScanNet \cite{bai2021pointdsc, Chen_2022_CVPR, zhang20233d}.

\subsection{Keypoint Matching on Outdoor}
\label{subsec:kpt_match}

Our method is equipped with several widely-used matchers, including the NN, OT and Mutual Nearest Neighbor (MNN) \cite{ao2021spinnet, huang2021predator}.
Recent studies \cite{zhang20233d, Chen_2022_CVPR} had shown that FCGF \cite{choy2019fully} exhibits little difference compared with FPFH \cite{rusu2009fast} on the KITTI dataset. Therefore, we choose FPFH and GEDI \cite{Poiesi2021} as the local geometric descriptors for evaluation.

\begin{table*}[t!]
	\renewcommand{\arraystretch}{1.2}
	\centering
	\renewcommand\tabcolsep{1.0pt}
	\caption{Keypoint matching comparison on KITTI.}
	\resizebox{1.0\linewidth}{!}{
		\begin{tabular}{c|cc|cc|cc|cc|cc|cc}
			\hline
			\multirow{2}[1]{*}{Matcher} & \multicolumn{6}{c|}{FPFH} & \multicolumn{6}{c}{GEDI}                                                                                                                                                                                                                                              \\
			                            & \multicolumn{2}{c}{easy}  & \multicolumn{2}{c}{medium} & \multicolumn{2}{c|}{hard} & \multicolumn{2}{c}{easy} & \multicolumn{2}{c}{medium} & \multicolumn{2}{c}{hard}                                                                                                                             \\
			\hline
			                            & IN                        & IR($\%$)                   & IN                        & IR($\%$)                 & IN                         & IR($\%$)                 & IN                  & IR($\%$)          & IN                  & IR($\%$)          & IN                 & IR($\%$)         \\
			\hline
			MNN                         & 81.37                     & 8.81                       & 20.85                     & 2.50                     & 6.64                       & 0.81                     & 457.70              & \underline{34.16} & 156.46              & 13.45             & 51.55              & 4.72             \\
			NN                          & 241.28                    & 4.83                       & 72.75                     & 1.45                     & 25.11                      & 0.50                     & 1024.21             & 20.48             & 376.06              & 7.52              & 129.33             & 2.59             \\
			OT                          & 228.98                    & 4.58                       & 73.52                     & 1.47                     & 25.10                      & 0.50                     & 958.01              & 19.16             & 354.38              & 7.09              & 120.98             & 2.42             \\
			\hline
			MNN+Ours                    & 450.27                    & \textbf{38.10}             & 181.47                    & \textbf{18.41}           & 61.33                      & \textbf{7.56}            & 868.35              & \textbf{60.41}    & 389.18              & \textbf{34.33}    & 140.77             & \textbf{14.10}   \\
			NN+Ours                     & \textbf{1246.10}          & \underline{20.33}          & \underline{556.59}        & 8.69                     & \textbf{231.35}         & 4.01                     & \textbf{2046.21}    & 33.99             & \underline{1020.55} & 16.40             & \underline{414.38} & 6.35             \\
			OT+Ours                     & \underline{1240.34}       & 20.01                      & \textbf{580.45}           & \underline{9.13}         & \underline{229.59}            & \underline{4.64}         & \underline{2031.18} & 33.13             & \textbf{1040.10}    & \underline{17.10} & \textbf{444.18}    & \underline{7.21} \\
			\hline
		\end{tabular}
	}
	\label{tab:tab12}
\end{table*}

\begin{figure}[t!]
    \centering
    \includegraphics[width=1.0\linewidth]{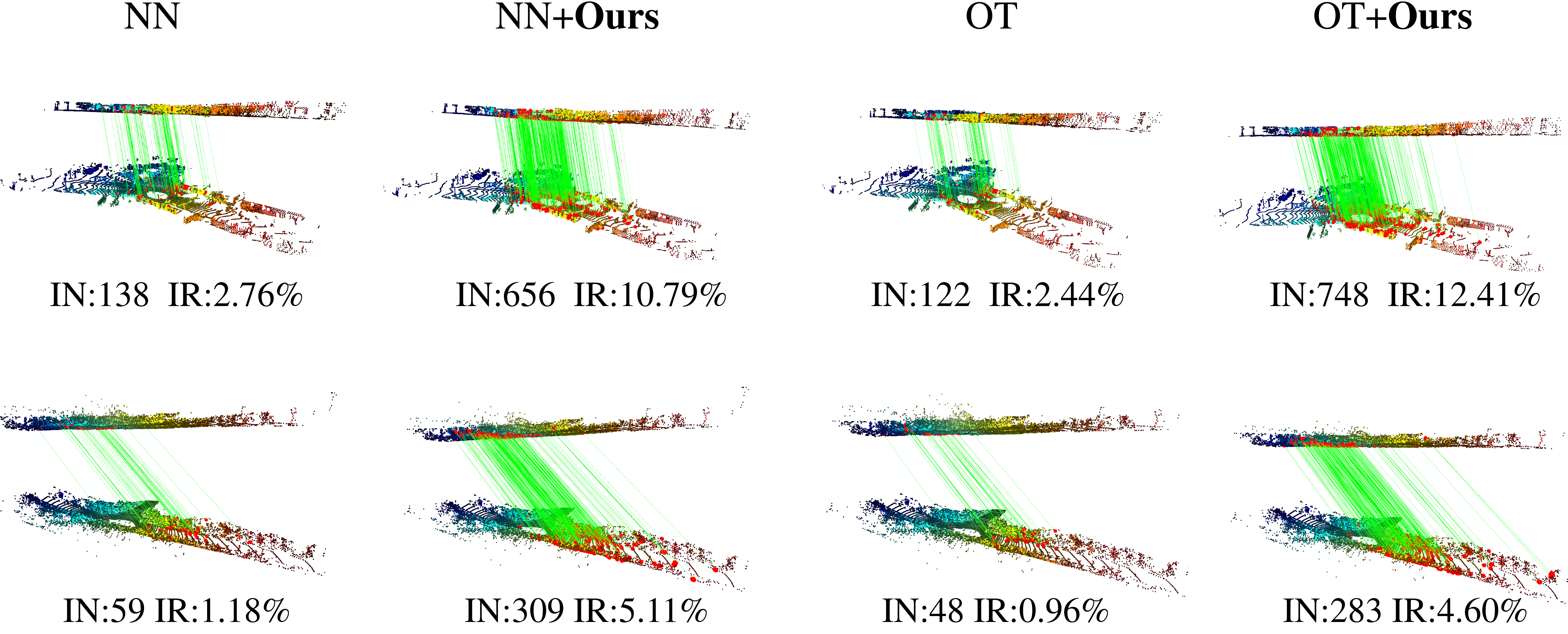}
    \caption{
        {
                Qualitative comparisons on the KITTI dataset of NN, NN+Ours, OT, and OT+Ours matchers using FPFH descriptor. The first and second rows depict examples from the easy (10-20m) and hard (20-30m) datasets, respectively. Inliers
                (residuals $<0.5m$)
                are indicated by green lines.} Best viewed on screen.
    }
    \label{fig:06}
\end{figure}

The correspondence evaluation results are shown in \cref{tab:tab12}.
Clearly, our method significantly enhances all baseline matchers. For example, when applying the FPFH and GEDI in the easy dataset,
the IN of NN matcher is $5$ and $2$ times improved after being equipped with our ML-SemReg.
The reasons are as follows:
(1) The Group Matching with Local-SS enhances the matcher's capability to perceive local-level semantic information, which can prevent inconsistent correspondences.
(2) The issue of missing matches between classes is mitigated by Group Matching, as it allows a single keypoint to be matched with multiple keypoints.
(3) SS-Consistency is achieved by the MM module using the scene-aware descriptor BMR-SS.
This reduces the
matching candidate set to
a refined subset, thus increasing the matching hit rate.
In addition,
{the findings highlight that in both the medium and hard datasets, the IR of NN+Ours with FPFH is higher than that of NN with GEDI.} This suggests that traditional descriptors equipped with our ML-SemReg can easily catch up or even surpass learning-based ones.
We show qualitative results in \cref{fig:06}. It is observed that the NN and OT matchers provide significantly more accurate inliers when equipped with ML-SemReg.

\subsection{Pairwise Registration on Outdoor}
In this section, we use correspondences created by NN and NN+Ours in \cref{subsec:kpt_match} to estimate the rigid transformation. We employ RANSAC  \cite{fischler1981random}, TeaserPlus  \cite{Yang20tro-teaser}, SC2PCR  \cite{Chen_2022_CVPR}, MAC  \cite{zhang20233d}, and PointDSC  \cite{bai2021pointdsc} as the registration estimators, in which we use RANSAC1M and RANSAC4M to represent the RANSAC with $1$ million and $4$ million iterations, respectively.
The registration results are reported in \cref{tab:RR_fpfh} for FPFH and GEDI, respectively.

\begin{table*}[t!]
	\centering
	\renewcommand{\arraystretch}{1.1}
	\renewcommand\tabcolsep{2.5pt}
	\caption{Registration results on the KITTI with FPFH and GEDI descriptors.}
	\resizebox{1.0\linewidth}{!}{
		\begin{tabular}{c|c|ccc|ccc|ccc}
			\hline
			\multirow{2}[1]{*}{Descriptor} & \multirow{2}[1]{*}{Estimator}       & \multicolumn{3}{c|}{easy} & \multicolumn{3}{c|}{medium} & \multicolumn{3}{c}{hard}                                                                                                                       \\
			                               &                                     & RE($^\circ$)              & TE(cm)                      & RR(\%)                   & RE($^\circ$)     & TE(cm)            & RR(\%)            & RE($^\circ$)     & TE(cm)            & RR(\%)            \\ \hline
			\multirow{12}[1]{*}{FPFH}      & RANSAC1M  \cite{fischler1981random} & 1.71                      & 33.17                       & 19.13                    & 1.79             & 37.48             & {1.25}            & -                & -                 & {0.00}            \\
			                               & RANSAC4M  \cite{fischler1981random} & 1.73                      & 33.20                       & {42.60}                  & 1.56             & 38.82             & {3.38}            & 1.15             & 50.36             & {0.20}            \\
			                               & TeaserPlus  \cite{Yang20tro-teaser} & 0.83                      & 15.93                       & 83.92                    & 1.47             & 29.34             & 32.83             & 1.39             & 36.36             & 2.85              \\
			                               & MAC  \cite{zhang20233d}             & 0.68                      & 14.64                       & 88.60                    & 1.48             & 32.26             & 35.50             & 1.89             & 31.46             & 1.43              \\
			                               & SC2PCR  \cite{Chen_2022_CVPR}       & 0.35                      & 10.15                       & 90.69                    & 0.82             & 22.45             & 63.17             & 1.01             & 31.53             & 8.56              \\
			                               & PointDSC  \cite{bai2021pointdsc}    & 0.35                      & 10.24                       & 89.64                    & 0.81             & 22.28             & 50.05             & 1.10             & 35.09             & 6.99              \\
			\cline{2-11}
			                               & RANSAC1M+Ours                       & 1.94                      & 31.06                       & 57.85                    & 2.10             & 36.18             & 26.23             & 2.26             & 37.99             & 6.93              \\
			                               & RANSAC4M+Ours                       & 1.91                      & 29.98                       & 58.71                    & 2.04             & 35.35             & 33.98             & 1.95             & 37.18             & 11.10             \\
			                               & TeaserPlus+Ours                     & 0.39                      & 10.01                       & 95.42                    & 0.87             & 17.91             & 85.49             & 1.34             & 27.49             & 55.99             \\
			                               & MAC+Ours                            & 0.44                      & 7.91                        & 96.43                    & 0.69             & 14.70             & 87.01             & 0.95             & 25.71             & 56.33             \\
			                               & SC2PCR+Ours                         & \underline{0.29}          & \underline{7.00}            & \underline{96.49}        & \underline{0.51} & \underline{12.61} & \textbf{91.01}    & \underline{0.70} & \underline{21.01} & \underline{67.82} \\
			                               & PointDSC+Ours                       & \textbf{0.24}             & \textbf{6.81}               & \textbf{96.61}           & \textbf{0.44}    & \textbf{12.31}    & \underline{90.11} & \textbf{0.69}    & \textbf{19.99}    & \textbf{68.13}    \\
			\hline
			\hline
			\multirow{12}[1]{*}{GEDI}      & RANSAC1M  \cite{fischler1981random} & 1.40                      & 24.91                       & {72.04}                  & 1.58             & 29.78             & {27.75}           & 1.07             & 36.22             & {1.43}            \\
			                               & RANSAC4M  \cite{fischler1981random} & 1.40                      & 25.34                       & {76.29}                  & 1.55             & 29.56             & {44.75}           & 1.61             & 30.89             & {4.89}            \\
			                               & TeaserPlus  \cite{Yang20tro-teaser} & 0.28                      & 7.74                        & 95.18                    & 0.59             & 15.51             & 88.58             & 0.91             & 26.54             & 51.71             \\
			                               & MAC  \cite{zhang20233d}             & 0.24                      & 6.39                        & 95.94                    & 0.49             & 13.05             & 89.50             & 0.93             & 27.51             & 40.53             \\
			                               & SC2PCR  \cite{Chen_2022_CVPR}       & \textbf{0.20}             & \underline{4.80}            & 96.32                    & 0.36             & 10.74             & 89.60             & 0.55             & 20.94             & 55.86             \\
			                               & PointDSC  \cite{bai2021pointdsc}    & \textbf{0.20}             & \textbf{4.77}               & 96.18                    & 0.37             & 10.79             & 91.69             & 0.53             & 20.50             & 50.50             \\
			\cline{2-11}
			                               & RANSAC1M+Ours                       & 1.61                      & 25.91                       & 79.55                    & 1.68             & 30.84             & 50.36             & 1.54             & 34.61             & 17.01             \\
			                               & RANSAC4M+Ours                       & 1.60                      & 26.30                       & 77.59                    & 1.91             & 31.88             & 59.48             & 1.79             & 36.21             & 24.34             \\
			                               & TeaserPlus+Ours                     & 0.25                      & 7.19                        & 97.50                    & 0.56             & 12.10             & 94.35             & 0.81             & 23.01             & 78.85             \\
			                               & MAC+Ours                            & \underline{0.22}          & 5.79                        & 97.62                    & 0.46             & 11.00             & 94.99             & 0.60             & 20.99             & 74.65             \\
			                               & SC2PCR+Ours                         & \textbf{0.20}             & 5.20                        & \textbf{98.10}           & \textbf{0.23}    & \textbf{9.71}     & \underline{96.23} & \underline{0.50} & \underline{17.81} & \underline{81.20} \\
			                               & PointDSC+Ours                       & \textbf{0.20}             & 5.20                        & \underline{97.91}        & \underline{0.33} & \underline{10.01} & \textbf{96.31}    & \textbf{0.44}    & \textbf{16.43}    & \textbf{82.01}    \\
			\hline
		\end{tabular}

	}
	\label{tab:RR_fpfh}
\end{table*}

\para{Enhancing Registration Recall.}
From \cref{tab:RR_fpfh}, we can draw the following conclusions:
(1) Our method can boost traditional RANSAC-type estimators by a large margin since it can effectively improve the IR of correspondences. The number of iterations required for RANSAC to correctly estimate the model increases exponentially as the IR decreases. In practical applications, the maximum number of iterations of RANSAC is usually limited to save time, which leads to its frequent failure when the IR is low.
(2) Our method can also improve the performance of SOTA estimators. The more difficult the registration, the more obvious the improvement. This is due to the fact that our method can simultaneously improve the IR and IN of initial correspondence.
On the one hand, a higher IR ensures better algorithm convergence.
On the other hand, a higher number of inliers improves the ability to identify
outliers \cite{zhang20233d, yang2023mutual}, thereby enhancing the quality of correspondences. For example, our approach enhances the SC2PCR and MAC by $59$ and $54$ percentage points, respectively, when utilizing the FPFH descriptor on the hard dataset.
(3) Our method shows good robustness in low-overlap scenarios, e.g.,
when employing SC2PCR, the RR of SC2PCR+Ours on the medium dataset reaches  a remarkable $91.01\%$, even surpassing the RR of SC2PCR on the easy dataset.

\para{Enhancing Handcrafted to Suppress Learning-based.}
Even without using any
data tranining
techniques, our method significantly improves the performance of FPFH (handcrafted) and is comparable to learning-based descriptors.
As shown in \cref{tab:tab12,tab:RR_fpfh}, e.g.,
{using NN+Ours+SC2PCR with FPFH setting yields IR=$8.69\%$ and RR=$91.01\%$, slightly outperforming NN+SC2PCR with GEDI descriptor in the medium case, where IR=$7.52\%$ and RR=$89.60\%$.}

\para{Enhancing Registration Accuracy.}
As shown in \cref{tab:RR_fpfh}, our algorithm shows higher registration accuracy in most scenarios.
The reasons may be twofold:
(1) ML-SemReg can generate correspondences with a higher IR, which is important for the convergence of registration estimators such as the MAC \cite{zhang20233d}.
(2) Our method enhances the creation of distant correspondences, which is more conducive to minimizing the transformation error.
Please refer to the visualization results in the supplementary material for details.
We can also see that although the TE of SC2PCR and PointDSC using GEDI descriptor on the easy dataset slightly increases after using ML-SemReg, our RR  is better.
This may be due to the fact that the quality of correspondences is already good enough in this easy case, causing our method to not improve the registration accuracy.

\subsection{Pairwise Registration On Indoor}

\begin{table}[t!]
	\centering
	\renewcommand{\arraystretch}{1.1}
	\renewcommand\tabcolsep{2.5pt}
	\caption{Results of correspondences and  registration on the ScanNet.}
	\resizebox{0.95\linewidth}{!}{
		\begin{tabular}{c|ccccc|ccccc}
			\hline
			\multirow{2}{*}{Registrator}        & \multicolumn{5}{c|}{40\% $\sim$ 50\%} & \multicolumn{5}{c}{50\%$\sim$60\%}                                                                                                                                                                                               \\
			                                    & IN                                    & IR(\%)                             & RE($^\circ$)     & TE(cm)           & RR(\%)            & IN                                  & IR(\%)                            & RE($^\circ$)     & TE(cm)           & RR(\%)            \\
			\hline
			RANSAC1M \cite{fischler1981random}  & \multirow{6}{*}{\underline{204.99}}   & \multirow{6}{*}{\underline{6.95}}  & 5.93             & 14.48            & 54.04             & \multirow{6}{*}{\underline{269.85}} & \multirow{6}{*}{\underline{9.31}} & 5.58             & 14.14            & 70.51             \\
			RANSAC4M \cite{fischler1981random}  &                                       &                                    & 6.02             & 14.86            & 59.58             &                                     &                                   & 5.68             & 14.20            & 71.33             \\
			TeaserPlus  \cite{Yang20tro-teaser} &                                       &                                    & 3.00             & 8.76             & 73.02             &                                     &                                   & 2.58             & 7.52             & 84.39             \\
			PointDSC  \cite{bai2021pointdsc}    &                                       &                                    & 3.13             & 9.78             & 73.11             &                                     &                                   & 3.18             & 8.01            & 76.31             \\
			MAC  \cite{zhang20233d}             &                                       &                                    & 5.28             & 13.92            & 46.57             &                                     &                                   & 4.93             & 12.57            & 61.43             \\
			SC2PCR  \cite{Chen_2022_CVPR}       &                                       &                                    & \underline{2.82} & {8.34} & 79.88             &                                     &                                   & \underline{2.36} & {7.00} & 87.55             \\
			\hline
			RANSAC1M+Ours                       & \multirow{6}{*}{\textbf{326.32}}      & \multirow{6}{*}{\textbf{9.32}}     & 6.31             & 15.71            & 64.31             & \multirow{6}{*}{\textbf{443.32}}    & \multirow{6}{*}{\textbf{12.03}}   & 6.01             & 15.61            & 74.88             \\
			RANSAC4M+Ours                       &                                       &                                    & 5.99             & 14.98            & 62.97             &                                     &                                   & 5.75             & 14.49            & 68.64             \\
			TeaserPlus+Ours                     &                                       &                                    & 3.84             & \underline{8.31}             & {82.33} &                                     &                                   & 3.53             & \underline{6.99}             & \underline{90.01} \\
			PointDSC+Ours                       &                                       &                                    & 3.15             & 9.01            & \underline{86.91}             &                                     &                                   & 3.13             & 7.43            & 89.24             \\
			MAC+Ours                            &                                       &                                    & 4.65             & 14.92            & 61.15             &                                     &                                   & 5.21             & 13.88            & 74.01             \\
			SC2PCR+Ours                         &                                       &                                    & \textbf{2.65}    & \textbf{8.12}    & \textbf{89.04}    &                                     &                                   & \textbf{2.12}    & \textbf{6.73}    & \textbf{92.78}    \\
			\hline
		\end{tabular}

	}
	\label{fig:fjdkasljfk}
\end{table}

The results of NN and NN+Ours with FPFH descriptor on the indoor ScanNet dataset are reported in \cref{fig:fjdkasljfk}.
Again, our pipelines significantly outperform the original methods on all evaluation metrics.
For instance, our method enhances the RR of SC2PCR by $9.16\%$ for overlap ratios ranging from $40\%$ to $50\%$, and by $5.23\%$ for overlap ratios ranging from $50\%$ to $60\%$.

\subsection{Robustness to Semenatic Segmentation Models}
To evaluate the robustness of ML-SemReg to semantic 
labels, we equipped it with five different semantic segmentation models (from 2019 to 2022). 
The correspondence evaluation results
are shown in \cref{tab:robustness_sem_model}.
We can clearly see that our method is stable across different models. This is because (1) the Local-SS employed by Group Matching exhibits robustness to segmentation noise, and (2) the landmark categories (such as car, pole, trunk) required for BMR-SS construction are easy-to-segment, thus the differences between different models are not significant.
We also provide more analyses experiments in supplementary material.

\begin{table}[t!]
    \centering
    \renewcommand{\arraystretch}{1.2}
    \renewcommand\tabcolsep{2.5pt}
    \caption{Correspondence evaluation results when using semantic labels from different semantic segmentation models on the three types of datasets. Our method demonstrates stability across different models.}
    \resizebox{0.7\linewidth}{!}{
        \begin{tabular}{c|c|cccccc}
            \hline
            \multirow{2}{*}{Method}                & \multirow{2}{*}{mIoU} & \multicolumn{2}{c}{easy} & \multicolumn{2}{c}{medium} & \multicolumn{2}{c}{hard}                                                                                    \\
                                                   &                       & \multicolumn{1}{c}{IN}   & \multicolumn{1}{c}{IR(\%)} & \multicolumn{1}{c}{IN}   & \multicolumn{1}{c}{IR(\%)} & \multicolumn{1}{c}{IN} & \multicolumn{1}{c}{IR(\%)} \\
            \hline
            RangeNet++(2019)\cite{milioto2019iros} & 52.2                  & 1877.45                  & 29.81                      & 873.51                   & 13.44                      & 321.11                 & 5.31                       \\
            KPConv(2019)\cite{thomas2019kpconv}    & 58.8                  & \textbf{2056.72}         & 32.85                      & \textbf{1075.59}         & 16.44                      & 425.34                 & 6.61                       \\
            Cylinder3D(2021)\cite{cylinder3d}      & 63.7                  & \underline{2054.04}      & 32.53                      & \underline{1052.47}      & \underline{16.54}          & 448.60                 & \textbf{6.88}              \\
            SPVNAS(2020)\cite{tang2020searching}   & 66.4                  & 2034.78                  & \underline{33.01}          & 1028.19                  & 16.19                      & 434.65                 & 6.40                       \\
            PTv2(2022)\cite{wu2022point}           & 70.3                  & 2029.30                  & 32.42                      & 1026.37                  & \textbf{16.55}             & \textbf{451.40}        & 6.71                       \\
            2DPASS(2022)\cite{yan20222dpass}       & 72.9                  & 2025.08                  & \textbf{33.03}             & 1037.57                  & 16.37                      & \underline{434.78}     & \underline{6.74}           \\
            \hline
        \end{tabular}
    }
    \label{tab:robustness_sem_model}
\end{table}

\subsection{Parameter Analysis Experiments}
For better understanding of our ML-SemReg, we conduct comprehensive experiments to study the parameter sensitivities,
including $N$ and $L$ of the BMR-SS, and $K$ of the binary selection function $\text{TL}^K(\cdot)$.
We employ the {NN+Ours with GEDI descriptor} on the three types of KITTI datasets for study.

\begin{figure}[tb]
    \centering
    \begin{subfigure}{0.8\linewidth}
        \includegraphics[width=\linewidth]{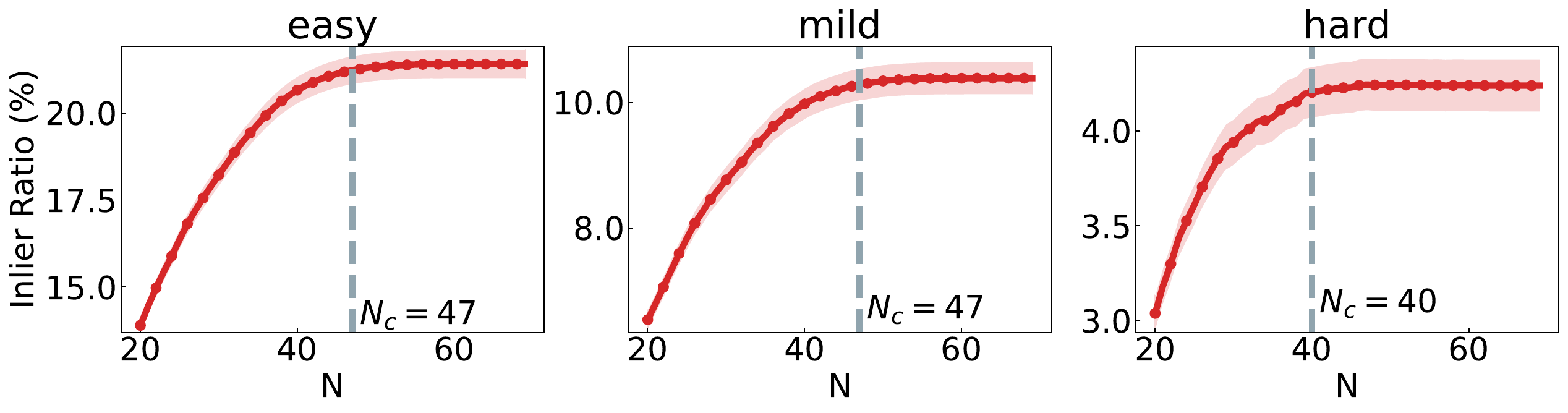}
        \caption{$L=2(m)$ and span with $\pm 0.05 \delta$.}
        \label{fig1_a}
    \end{subfigure}
    \vfill
    \begin{subfigure}{0.8\linewidth}
        \includegraphics[width=\linewidth]{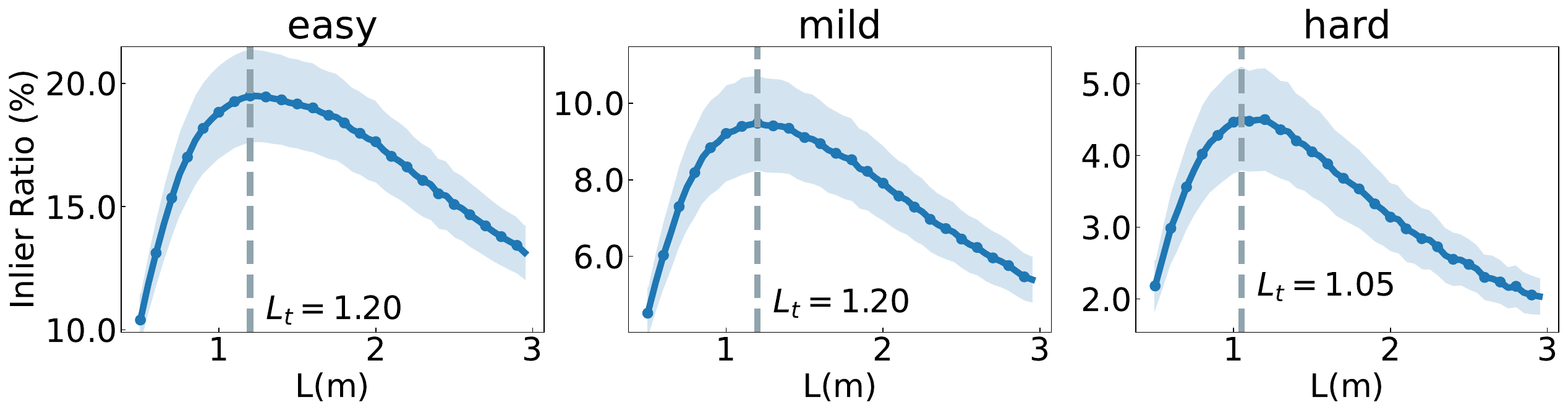}
        \caption{$N=30$ and span with $\pm 0.2 \delta$.}
        \label{fig1_b}
    \end{subfigure}
    \caption{
        Sensitivity of IR to $N$ and $L$. Each scatter is the mean of metrics, and $\delta$ is the standard deviation of all samples.
    }
    \label{figggg}
\end{figure}

\para{Receptive Field and Precision of BMR-SS.}
\cref{figggg} illustrates our exploration of two key parameters of BMR-SS:
(1) The ring number, denoted as $N$; (2) The ring length, denoted as $L$ (m).
$N_c$ is defined as the first $N$ value at which IR exceeds 99\% of the maximum IR in \cref{fig1_a}
and $L_t$ corresponds to the maximum IR in \cref{fig1_b}.
As depicted in \cref{fig1_a}, IR gradually improves with the increase of $N$ as the expansion of perceptual field, since more relevant scene information can be encoded in BMR-SS.
We notice that when $N > N_c$, the perceptual field already encompasses almost all relevant semantic landmarks and thus only brings marginal improvements.
$N_c$ of hard dataset is lower than other two datasets.
This may caused by oversized receptive fields which contain some uncorrelated landmarks in non-overlapping parts, thus negatively affecting BMR-SS.
Moving to \cref{fig1_b}, IR initially improves with the increase of receptive field.
However, IR starts to decrease after surpassing $L_t$. This decrease may caused by large values of $L$, which result in a large number of landmarks within a ring that satisfy the SS-Consistency, thereby reducing localization accuracy.
Despite the noticeable changes in IR with the variations of $N$ and $L$, the optimal parameter settings remain stable, indicating the robustness of our method across diverse overlap scenarios.

\begin{table}[t!]
	\centering%
	\renewcommand{\arraystretch}{1.4}
	\scriptsize
	\renewcommand\tabcolsep{1.0pt}
	\caption{
		Analysis the effect of $K$ in $\text{TL}^K$ function.
		Bold indicates the maximum value, underline indicates the second maximum, and `-' indicates not using ML-SemReg.
	}
	\resizebox{0.9\linewidth}{!}{
		\begin{tabular}{c|c|c|cccccccccc}
			\hline
			                      & K      & -      & 1      & 2                   & 3                  & 4       & 5      & 6      & 7      & 8      & 9      & 10     \\
			\hline
			\multirow{2}{*}{FPFH} & IN     & 172.01 & 129.17 & \textbf{821.11}     & \underline{666.33} & 489.48  & 323.80 & 289.17 & 233.82 & 200.29 & 186.12 & 167.21 \\
			                      & IR(\%) & 2.49   & 3.30   & \textbf{13.10}      & \underline{10.77}  & 7.85    & 5.99   & 4.18   & 3.56   & 3.26   & 2.98   & 2.33   \\
			\hline
			\multirow{2}{*}{GEDI} & IN     & 627.88 & 557.73 & \underline{1085.01} & \textbf{1151.11}   & 1025.11 & 882.98 & 791.44 & 741.32 & 682.88 & 665.33 & 636.12 \\
			                      & IR(\%) & 11.01  & 12.44  & \underline{17.98}   & \textbf{18.68}     & 16.68   & 15.11  & 13.33  & 12.11  & 11.44  & 10.88  & 10.38  \\
			\hline
		\end{tabular}
	}
	\label{table:Kparameter}
\end{table}

\para{Balance of BMR-SS and Descriptor.}
The parameter $K$ in the function $\text{TL}^K(\cdot)$ is used to retain the top-$K$ maximal values in each row of the scene similarity matrix.
\cref{table:Kparameter} studies the impact of parameter $K$ on matching performance {using NN with FPFH/GEDI descriptors}.
We start with $K=1$ and gradually increase it to $10$.
When $K=1$, MM over-relies on the BMR-SS description, so our method performs even worse than the original matcher (NN).
Our method achieves the best performance when $K=2$ or $3$.
The results also show that any value of $K>1$ outperforms the raw matchers.
As $K$ increases, the influence of the MM diminishes, causing the matcher to rely more and more on unstable local geometric features and leading to degradation.
This suggests that local geometric features and BMR-SS  can be balanced by $K$.

\subsection{Ablation Study}
We systematically evaluate the effectiveness of all components utilizing NN matcher, SC2PCR estimator, and FPFH/GEDI descriptors.
We progressively integrate the GM and MM modules. Furthermore, we perform a comparative analysis between
\orangetext{Matching in the Same Category (MSC) and matching in the matching group based on the LS-Consistency (i.e., GM).}
The results are shown in \cref{tab:ablation}, from which we can draw the following conclusions:
(1) Performance drops after using the MSC (comparing row 1 with 0), since the estimated semantic labels are inaccurate (semantic noise), producing new outliers and causing inlier missing.
(2) The LS-Consistency overcoms the semantic noise problem and the GM module reduces the search space of correspondence establishment (comparing row 2 with 1 and 0), thus improving the IN and RR.
(3) The MM module can significantly improve matching performance (comparing row 3 with 0), i.e., the RR is improved from $54.33\%$ to $83.44\%$ when employing the FPFH descriptor, and improved from $83.77\%$ to $91.33\%$ when employing GEDI.
This improvement stems from its good scene awareness ability, which is complementary to local geometric similarity.
(4) The combination of GM and MM modules achieves the best performance since they ensure multi-level semantic consistency, allowing them to overcome both inter-class and intra-class mismatching problems, thereby further improving the overall accuracy.

\begin{table}[t!]
	\centering
	\renewcommand{\arraystretch}{1.2}
	\scriptsize
	\renewcommand\tabcolsep{2.5pt}
	\caption{Ablation study on KITTI
			when leveraging NN matcher, SC2PCR estimator, and FPFH/GEDI descriptors.
		\textbf{MSC:} Match in the Same Category.
		\textbf{GM:} Group Matching.
		\textbf{MM:} Mask Matching.
	}
	\resizebox{0.8\linewidth}{!}{
		\begin{tabular}{cc|cc|c|cccccc}
			\hline
			                      &      & MSC                      & GM        & MM        & IN                    & IR(\%)              & RE($^\circ$)       & TE(cm)              & RR(\%)              \\ \hline
			\multirow{5}{*}{FPFH} & 0)   &                          &            &            & 115.36                & 2.36
			                      & 0.71 & 21.44                    & 54.33                                                                                                                                  \\
			                      & 1)   & \checkmark               &            &            & 74.30                 & 1.51
			                      & 0.81 & 25.00                    & 52.56                                                                                                                                  \\
			                      & 2)   &                          & \checkmark &            & 161.74                & 2.60                & 0.63               & 20.31               & 60.90               \\
			                      & 3)   &                          &            & \checkmark & {\underline {550.20}} & \textbf{11.01}      & {\underline{0.52}} & {\underline{13.55}} & {\underline{83.44}} \\
			                      & 4)   &                          & \checkmark & \checkmark & \textbf{681.01}       & {\underline{10.12}} & \textbf{0.44}      & \textbf{13.33}      & \textbf{85.01}      \\ \hline
			\hline
			\multirow{5}{*}{GEDI} & 0)   &                          &            &            & 508.41                & 11.01               & 0.40               & 12.41               & 83.77               \\
			                      & 1)   & \checkmark               &            &            & 287.82                & 5.66                & 0.53               & 13.33               & 82.00               \\
			                      & 2)   &                          & \checkmark &            & 615.73                & 9.89                & \underline{0.34}   & 12.25               & 83.99               \\
			                      & 3)   &                          &            & \checkmark & {\underline{1014.82}} & \textbf{20.50}      & \underline{0.34}   & {\underline{11.56}} & {\underline{91.33}} \\
			                      & 4)   &                          & \checkmark & \checkmark & \textbf{1143.41}      & {\underline{18.80}} & \textbf{0.32}      & \textbf{11.31}      & \textbf{91.88}      \\

			\hline
		\end{tabular}
	}
	\label{tab:ablation}
\end{table}

\section{Conclusion}
\label{sec:conclusion}
In this work, we presented ML-SemReg, a plug-and-play framework that is designed to enhance correspondence-based PCR by leveraging multi-level semantic consistency. To address inter-class mismatching, we introduced a novel Group Matching module, which restricts matching to groups that inherently satisfy Local Semantic Consistency. To suppress intra-class mismatching, we proposed a  Mask Matching module  based on our scene-aware Binary Multi-Ring Semantic Signature descriptor. Extensive experiments on indoor and outdoor datasets demonstrated that our method significantly outperforms state-of-the-art approaches and exhibits robustness to semantic labels.  

\section*{Acknowledgements}
This work was supported by the the Wuhan university-Huawei Geoinformatics Innovation Laboratory Open Fund under Grant TC20210901025-2023-06.

%
%
\bibliographystyle{splncs04}
\bibliography{main}

\begin{thebibliography}{10}
\providecommand{\url}[1]{\texttt{#1}}
\providecommand{\urlprefix}{URL }
\providecommand{\doi}[1]{https://doi.org/#1}

\bibitem{aiger20084}
Aiger, D., Mitra, N.J., Cohen-Or, D.: 4-points congruent sets for robust
  pairwise surface registration. In: ACM SIGGRAPH 2008 papers, pp. 1--10 (2008)

\bibitem{ao2021spinnet}
Ao, S., Hu, Q., Yang, B., Markham, A., Guo, Y.: Spinnet: Learning a general
  surface descriptor for 3d point cloud registration. In: Proceedings of the
  IEEE/CVF conference on computer vision and pattern recognition. pp.
  11753--11762 (2021)

\bibitem{arce2023padloc}
Arce, J., V{\"o}disch, N., Cattaneo, D., Burgard, W., Valada, A.: Padloc:
  Lidar-based deep loop closure detection and registration using panoptic
  attention. IEEE Robotics and Automation Letters  \textbf{8}(3),  1319--1326
  (2023)

\bibitem{azuma1997survey}
Azuma, R.T.: A survey of augmented reality. Presence: teleoperators \& virtual
  environments  \textbf{6}(4),  355--385 (1997)

\bibitem{besl1992method}
Besl, P.J., McKay, N.D.: Method for registration of 3-d shapes. In: Sensor
  fusion IV: control paradigms and data structures. vol.~1611, pp. 586--606.
  International Society for Optics and Photonics (1992)

\bibitem{bouaziz2013sparse}
Bouaziz, S., Tagliasacchi, A., Pauly, M.: Sparse iterative closest point. In:
  Computer graphics forum. vol.~32, pp. 113--123. Wiley Online Library (2013)

\bibitem{chen20073d}
Chen, H., Bhanu, B.: 3d free-form object recognition in range images using
  local surface patches. Pattern Recognition Letters  \textbf{28}(10),
  1252--1262 (2007)

\bibitem{chen2019iros}
Chen, X., Milioto, A., Palazzolo, E., Giguère, P., Behley, J., Stachniss, C.:
  {SuMa++: Efficient LiDAR-based Semantic SLAM}. In: Proceedings of the
  IEEE/RSJ Int. Conf. on Intelligent Robots and Systems (IROS). pp. 4530--4537
  (2019),
  \url{https://www.ipb.uni-bonn.de/wp-content/papercite-data/pdf/chen2019iros.pdf}

\bibitem{Chen_2022_CVPR}
Chen, Z., Sun, K., Yang, F., Tao, W.: Sc2-pcr: A second order spatial
  compatibility for efficient and robust point cloud registration. In:
  Proceedings of the IEEE/CVF Conference on Computer Vision and Pattern
  Recognition (CVPR). pp. 13221--13231 (June 2022)

\bibitem{Choi_2015_CVPR}
Choi, S., Zhou, Q.Y., Koltun, V.: Robust reconstruction of indoor scenes. In:
  Proceedings of the IEEE Conference on Computer Vision and Pattern Recognition
  (CVPR) (June 2015)

\bibitem{choy2020deep}
Choy, C., Dong, W., Koltun, V.: Deep global registration. In: Proceedings of
  the IEEE/CVF conference on computer vision and pattern recognition. pp.
  2514--2523 (2020)

\bibitem{choy2019fully}
Choy, C., Park, J., Koltun, V.: Fully convolutional geometric features. In:
  Proceedings of the IEEE International Conference on Computer Vision. pp.
  8958--8966 (2019)

\bibitem{dai2017scannet}
Dai, A., Chang, A.X., Savva, M., Halber, M., Funkhouser, T., Nie{\ss}ner, M.:
  Scannet: Richly-annotated 3d reconstructions of indoor scenes. In:
  Proceedings of the IEEE conference on computer vision and pattern
  recognition. pp. 5828--5839 (2017)

\bibitem{deng2018ppf}
Deng, H., Birdal, T., Ilic, S.: Ppf-foldnet: Unsupervised learning of rotation
  invariant 3d local descriptors. In: Proceedings of the European conference on
  computer vision (ECCV). pp. 602--618 (2018)

\bibitem{deng2018ppfnet}
Deng, H., Birdal, T., Ilic, S.: Ppfnet: Global context aware local features for
  robust 3d point matching. In: Proceedings of the IEEE conference on computer
  vision and pattern recognition. pp. 195--205 (2018)

\bibitem{durrant2006simultaneous}
Durrant-Whyte, H., Bailey, T.: Simultaneous localization and mapping: part i.
  IEEE robotics \& automation magazine  \textbf{13}(2),  99--110 (2006)

\bibitem{fischler1981random}
Fischler, M.A., Bolles, R.C.: Random sample consensus: a paradigm for model
  fitting with applications to image analysis and automated cartography.
  Communications of the ACM  \textbf{24}(6),  381--395 (1981)

\bibitem{frome2004recognizing}
Frome, A., Huber, D., Kolluri, R., B{\"u}low, T., Malik, J.: Recognizing
  objects in range data using regional point descriptors. In: European
  conference on computer vision. pp. 224--237. Springer (2004)

\bibitem{Geiger2012CVPR}
Geiger, A., Lenz, P., Urtasun, R.: Are we ready for autonomous driving? the
  kitti vision benchmark suite. In: Conference on Computer Vision and Pattern
  Recognition (CVPR) (2012)

\bibitem{gong2021boundary}
Gong, J., Xu, J., Tan, X., Zhou, J., Qu, Y., Xie, Y., Ma, L.: Boundary-aware
  geometric encoding for semantic segmentation of point clouds. In: Proceedings
  of the AAAI Conference on Artificial Intelligence. vol.~35, pp. 1424--1432
  (2021)

\bibitem{guo2015novel}
Guo, Y., Sohel, F., Bennamoun, M., Wan, J., Lu, M.: A novel local surface
  feature for 3d object recognition under clutter and occlusion. Information
  Sciences  \textbf{293},  196--213 (2015)

\bibitem{guo2013rops}
Guo, Y., Sohel, F.A., Bennamoun, M., Wan, J., Lu, M.: Rops: A local feature
  descriptor for 3d rigid objects based on rotational projection statistics.
  In: 2013 1st International Conference on Communications, Signal Processing,
  and their Applications (ICCSPA). pp.~1--6. IEEE (2013)

\bibitem{he2021vi}
He, Y., Ma, L., Jiang, Z., Tang, Y., Xing, G.: Vi-eye: Semantic-based 3d point
  cloud registration for infrastructure-assisted autonomous driving. In:
  Proceedings of the 27th Annual International Conference on Mobile Computing
  and Networking. pp. 573--586 (2021)

\bibitem{huang2021predator}
Huang, S., Gojcic, Z., Usvyatsov, M., Wieser, A., Schindler, K.: Predator:
  Registration of 3d point clouds with low overlap. In: Proceedings of the
  IEEE/CVF Conference on computer vision and pattern recognition. pp.
  4267--4276 (2021)

\bibitem{johnson1999using}
Johnson, A.E., Hebert, M.: Using spin images for efficient object recognition
  in cluttered 3d scenes. IEEE Transactions on pattern analysis and machine
  intelligence  \textbf{21}(5),  433--449 (1999)

\bibitem{lepard2021}
Li, Y., Harada, T.: Lepard: Learning partial point cloud matching in rigid and
  deformable scenes. IEEE/CVF Conference on Computer Vision and Pattern
  Recognition (CVPR)  (2022)

\bibitem{sift2012}
Lindeberg, T.: Scale invariant feature transform  (2012)

\bibitem{liu2022sarnet}
Liu, C., Guo, J., Yan, D.M., Liang, Z., Zhang, X., Cheng, Z.: Sarnet: Semantic
  augmented registration of large-scale urban point clouds. arXiv preprint
  arXiv:2206.13117  (2022)

\bibitem{liu2024deep}
Liu, S., Wang, T., Zhang, Y., Zhou, R., Li, L., Dai, C., Zhang, Y., Wang, L.,
  Wang, H.: Deep semantic graph matching for large-scale outdoor point cloud
  registration. IEEE Transactions on Geoscience and Remote Sensing  (2024)

\bibitem{milioto2019iros}
Milioto, A., Vizzo, I., Behley, J., Stachniss, C.: {RangeNet++: Fast and
  Accurate LiDAR Semantic Segmentation}. In: IEEE/RSJ Intl.~Conf.~on
  Intelligent Robots and Systems (IROS) (2019)

\bibitem{pais20203dregnet}
Pais, G.D., Ramalingam, S., Govindu, V.M., Nascimento, J.C., Chellappa, R.,
  Miraldo, P.: 3dregnet: A deep neural network for 3d point registration. In:
  Proceedings of the IEEE/CVF conference on computer vision and pattern
  recognition. pp. 7193--7203 (2020)

\bibitem{Poiesi2021}
Poiesi, F., Boscaini, D.: Learning general and distinctive 3d local deep
  descriptors for point cloud registration. In: IEEE Trans. on Pattern Analysis
  and Machine Intelligence ((early access) 2022)

\bibitem{qiao2023g3reg}
Qiao, Z., Yu, Z., Jiang, B., Yin, H., Shen, S.: G3reg: Pyramid graph-based
  global registration using gaussian ellipsoid model. arXiv preprint
  arXiv:2308.11573  (2023)

\bibitem{qin2022geometric}
Qin, Z., Yu, H., Wang, C., Guo, Y., Peng, Y., Xu, K.: Geometric transformer for
  fast and robust point cloud registration. In: Proceedings of the IEEE/CVF
  Conference on Computer Vision and Pattern Recognition (CVPR). pp.
  11143--11152 (June 2022)

\bibitem{rusinkiewicz2001efficient}
Rusinkiewicz, S., Levoy, M.: Efficient variants of the icp algorithm. In:
  Proceedings third international conference on 3-D digital imaging and
  modeling. pp. 145--152. IEEE (2001)

\bibitem{rusu2009fast}
Rusu, R.B., Blodow, N., Beetz, M.: Fast point feature histograms (fpfh) for 3d
  registration. In: 2009 IEEE international conference on robotics and
  automation. pp. 3212--3217. IEEE (2009)

\bibitem{rusu2008aligning}
Rusu, R.B., Blodow, N., Marton, Z.C., Beetz, M.: Aligning point cloud views
  using persistent feature histograms. In: 2008 IEEE/RSJ international
  conference on intelligent robots and systems. pp. 3384--3391. IEEE (2008)

\bibitem{segal2009generalized}
Segal, A., Haehnel, D., Thrun, S.: Generalized-icp. In: Robotics: science and
  systems. vol.~2, p.~435. Seattle, WA (2009)

\bibitem{tang2020searching}
Tang, H., Liu, Z., Zhao, S., Lin, Y., Lin, J., Wang, H., Han, S.: Searching
  efficient 3d architectures with sparse point-voxel convolution. In: European
  Conference on Computer Vision (2020)

\bibitem{tang2022contrastive}
Tang, L., Zhan, Y., Chen, Z., Yu, B., Tao, D.: Contrastive boundary learning
  for point cloud segmentation. In: Proceedings of the IEEE/CVF Conference on
  Computer Vision and Pattern Recognition. pp. 8489--8499 (2022)

\bibitem{thomas2019kpconv}
Thomas, H., Qi, C.R., Deschaud, J.E., Marcotegui, B., Goulette, F., Guibas,
  L.J.: Kpconv: Flexible and deformable convolution for point clouds. In:
  Proceedings of the IEEE/CVF international conference on computer vision. pp.
  6411--6420 (2019)

\bibitem{tombari2010unique}
Tombari, F., Salti, S., Di~Stefano, L.: Unique shape context for 3d data
  description. In: Proceedings of the ACM workshop on 3D object retrieval. pp.
  57--62 (2010)

\bibitem{DBLP:conf/dicta/TruongGIS19}
Truong, G., Gilani, S.Z., Islam, S.M.S., Suter, D.: Fast point cloud
  registration using semantic segmentation. In: 2019 Digital Image Computing:
  Techniques and Applications, {DICTA} 2019, Perth, Australia, December 2-4,
  2019. pp.~1--8 (2019). \doi{10.1109/DICTA47822.2019.8945870},
  \url{https://doi.org/10.1109/DICTA47822.2019.8945870}

\bibitem{wang2022you}
Wang, H., Liu, Y., Dong, Z., Wang, W.: You only hypothesize once: Point cloud
  registration with rotation-equivariant descriptors. In: Proceedings of the
  30th ACM International Conference on Multimedia. pp. 1630--1641 (2022)

\bibitem{wu2022point}
Wu, X., Lao, Y., Jiang, L., Liu, X., Zhao, H.: Point transformer v2: Grouped
  vector attention and partition-based pooling. In: NeurIPS (2022)

\bibitem{bai2020d3feat}
Xuyang~Bai, Zixin~Luo, L.Z.H.F.L.Q., Tai, C.L.: D3feat: Joint learning of dense
  detection and description of 3d local features. arXiv:2003.03164 [cs.CV]
  (2020)

\bibitem{bai2021pointdsc}
Xuyang~Bai, Zixin~Luo, L.Z.H.C.L.L.Z.H.H.F., Tai, C.L.: {PointDSC}: {R}obust
  {P}oint {C}loud {R}egistration using {D}eep {S}patial {C}onsistency. CVPR
  (2021)

\bibitem{yan20222dpass}
Yan, X., Gao, J., Zheng, C., Zheng, C., Zhang, R., Cui, S., Li, Z.: 2dpass: 2d
  priors assisted semantic segmentation on lidar point clouds. In: European
  Conference on Computer Vision. pp. 677--695. Springer (2022)

\bibitem{Yang20tro-teaser}
Yang, H., Shi, J., Carlone, L.: {TEASER: Fast and Certifiable Point Cloud
  Registration}. {IEEE} Trans. Robotics  (2020)

\bibitem{yang2023mutual}
Yang, J., Zhang, X., Fan, S., Ren, C., Zhang, Y.: Mutual voting for ranking 3d
  correspondences. IEEE Transactions on Pattern Analysis and Machine
  Intelligence  (2023)

\bibitem{yang2018foldingnet}
Yang, Y., Feng, C., Shen, Y., Tian, D.: Foldingnet: Point cloud auto-encoder
  via deep grid deformation. In: Proceedings of the IEEE Conference on Computer
  Vision and Pattern Recognition. pp. 206--215 (2018)

\bibitem{yew2022regtr}
Yew, Z.J., Lee, G.h.: Regtr: End-to-end point cloud correspondences with
  transformers. In: CVPR (2022)

\bibitem{yin2023segregator}
Yin, P., Yuan, S., Cao, H., Ji, X., Zhang, S., Xie, L.: Segregator: Global
  point cloud registration with semantic and geometric cues. arXiv preprint
  arXiv:2301.07425  (2023)

\bibitem{yu2021cofinet}
Yu, H., Li, F., Saleh, M., Busam, B., Ilic, S.: Cofinet: Reliable
  coarse-to-fine correspondences for robust pointcloud registration. Advances
  in Neural Information Processing Systems  \textbf{34},  23872--23884 (2021)

\bibitem{yu2023rotation}
Yu, H., Qin, Z., Hou, J., Saleh, M., Li, D., Busam, B., Ilic, S.:
  Rotation-invariant transformer for point cloud matching. In: Proceedings of
  the IEEE/CVF Conference on Computer Vision and Pattern Recognition. pp.
  5384--5393 (2023)

\bibitem{DBLP:journals/ral/ZaganidisSDC18}
Zaganidis, A., Sun, L., Duckett, T., Cielniak, G.: Integrating deep semantic
  segmentation into 3-d point cloud registration. {IEEE} Robotics Autom. Lett.
  \textbf{3}(4),  2942--2949 (2018)

\bibitem{zaharescu2009surface}
Zaharescu, A., Boyer, E., Varanasi, K., Horaud, R.: Surface feature detection
  and description with applications to mesh matching. In: 2009 IEEE Conference
  on Computer Vision and Pattern Recognition. pp. 373--380. IEEE (2009)

\bibitem{zeng20173dmatch}
Zeng, A., Song, S., Nie{\ss}ner, M., Fisher, M., Xiao, J., Funkhouser, T.:
  3dmatch: Learning local geometric descriptors from rgb-d reconstructions. In:
  Proceedings of the IEEE conference on computer vision and pattern
  recognition. pp. 1802--1811 (2017)

\bibitem{zhang20233d}
Zhang, X., Yang, J., Zhang, S., Zhang, Y.: 3d registration with maximal
  cliques. In: Proceedings of the IEEE/CVF Conference on Computer Vision and
  Pattern Recognition. pp. 17745--17754 (2023)

\bibitem{qiao_pagor}
Zhijian~Qiao, Zehuan~Yu, H.Y., Shen, S.: Pyramid semantic graph-based global
  point cloud registration with low overlap. In: 2023 IEEE/RSJ International
  Conference on Intelligent Robots and Systems (IROS) (2023)

\bibitem{Zhou2018open3d}
Zhou, Q.Y., Park, J., Koltun, V.: {Open3D}: {A} modern library for {3D} data
  processing. arXiv:1801.09847  (2018)

\bibitem{cylinder3d}
Zhu, X., Zhou, H., Wang, T., Hong, F., Ma, Y., Li, W., Li, H., Lin, D.:
  Cylindrical and asymmetrical 3d convolution networks for lidar segmentation.
  In: IEEE Conference on Computer Vision and Pattern Recognition. pp.
  9939--9948 (2021)

\end{thebibliography}

\appendix
\clearpage

\noindent{\huge\bfseries APPENDIX\par}

\section{Explanation of LS-Consistency in the Group Matching Module}

\begin{figure}[th!]
	\centering
	\includegraphics[width=.7\linewidth]{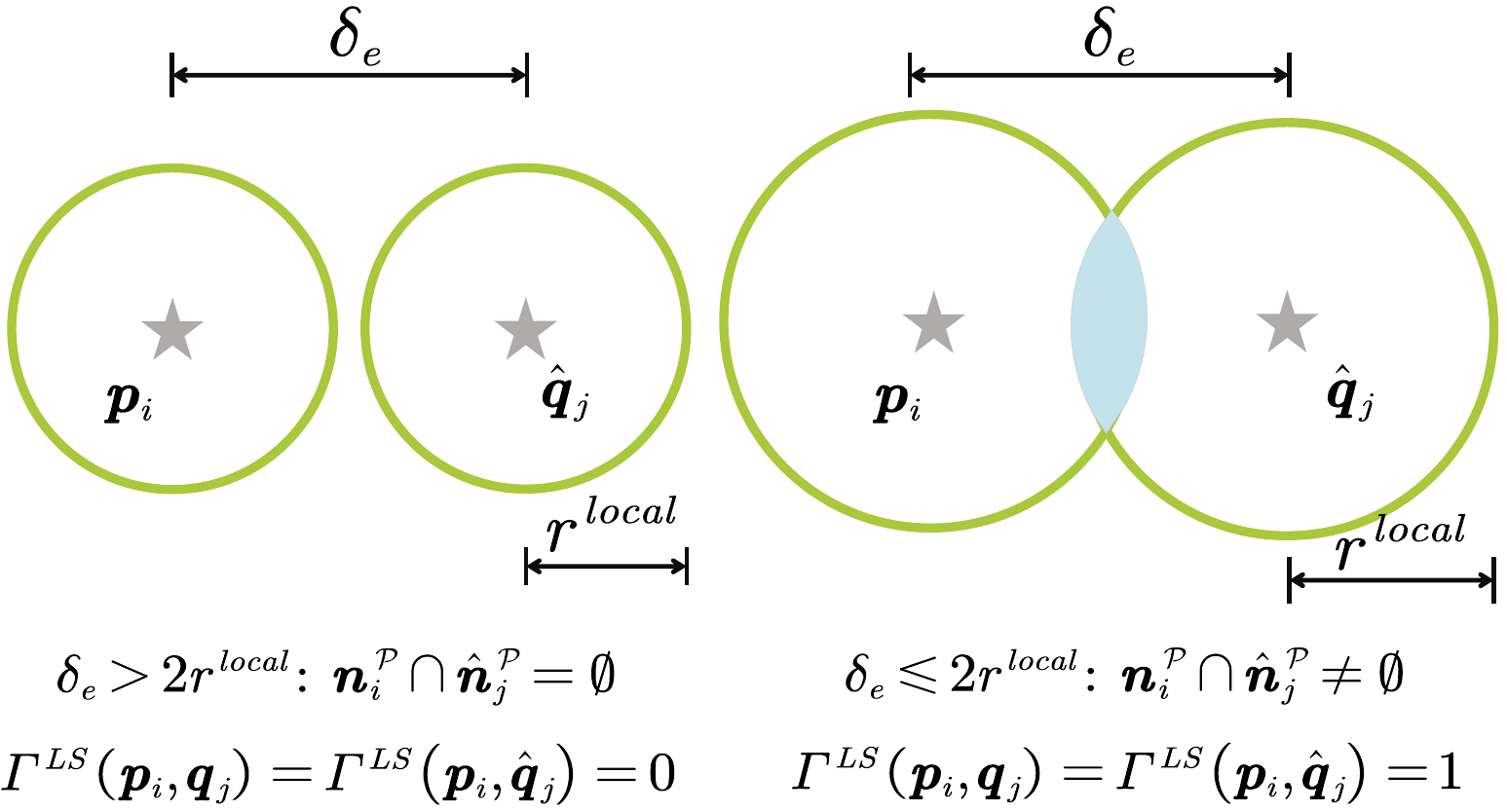}
	\caption{
        Visualization of LS-Consistency with varying neighborhood search radii $r^{local}$.
        The green circles around each keypoint represent the query region of Local-SS.
        (1) left: The neighborhood search radius is too small to form an intersection set, thereby not satisfying LS-Consistency. (2) right: When $r^{local} \ge \frac{\delta_e}{2}$, then these two points satisfy LS-Consistency as the intersection of their search circles is non-empty.
    }
	\label{fig:vis_nss}
\end{figure}

In this section, we proof the proposition in the Group Matching (\cref{subsec:group_matching}) module: "When the search radius $r^{local}$ is large enough, one correspondence can be classified as inter-class mismatching (outlier) if two keypoints cannot satisfy the LS-Consistency."

\begin{proof}
    Given a correspondence $(\boldsymbol{p}_{i}, \boldsymbol{q}_{j})$ and the ground truth transformation $(\mathbf{R}, \boldsymbol{t})$, the matching residual of $(\boldsymbol{p}_{i}, \boldsymbol{q}_{j})$ can be calculated as $\delta_{e}  = \Vert \boldsymbol{p}_{i} - \boldsymbol{\hat{q}}_j  \Vert_{2}$, where $\boldsymbol{\hat{q}}_j = \mathbf{R} \boldsymbol{q}_{j} + \boldsymbol{t}$ is the reprojection of $\boldsymbol{q}_j$.
    Then, we can classify whether the correspondence $(\boldsymbol{p}_{i}, \boldsymbol{q}_{j})$ is an inlier by:
    \begin{equation}
            \varOmega \left( \boldsymbol{p}_i,\boldsymbol{q}_j \right) =\begin{cases}
        1&		\text{if } \delta _e\le \tau _e\\
        0&		\text{if } \delta _e > \tau _e
    \end{cases}, 
    \label{vgfjkd}
    \end{equation}
    where $\varOmega \left( \boldsymbol{p}_i,\boldsymbol{q}_j \right)=1$ indicates that $(\boldsymbol{p}_i,\boldsymbol{q}_j)$ is an inlier; otherwise, it is an outlier. $\tau_{e}$ is an inlier threshold

    Let $\boldsymbol{\hat{n}}_{j}^{\mathcal{P}}$ represents the Local-SS of $\hat{\boldsymbol{q}}_j$.
    {Note that $\boldsymbol{\hat{q}}_j$ is reprojected into source point cloud, thus $\boldsymbol{\hat{n}}_{j}^{\mathcal{P}}$ is calculated via a local circular region of source point cloud centered at $\boldsymbol{\hat{q}}_j$.}
    Generally, we have $\boldsymbol{\hat{n}}_{j}^{\mathcal{P}}=\boldsymbol{n}_j^{\mathcal{Q}}$, thus:
    \begin{equation}
        \varGamma ^{NS}\left( \boldsymbol{p}_i,\boldsymbol{q}_j \right) =\varGamma ^{NS}\left( \boldsymbol{p}_i,\boldsymbol{\hat{q}}_j \right).
    \end{equation}
    where $\varGamma^{NS}$ indicates whether two points satisfy the LS-Consistency.
    Let $r^{local} > 0$ be the search radius for constructing the Local Semantic Signature (Local-SS).
    {\cref{fig:vis_nss} demonstrates that $\boldsymbol{p}_i$ and $\boldsymbol{q}_j$ (also $\hat{\boldsymbol{q}}_j$) satisfy LS-Consistency if $\delta_e \le 2r^{local}$:}
    \begin{equation}
        \delta_e \le 2r^{local} \rightarrow \varGamma^{NS}\left(\boldsymbol{p}_i,\boldsymbol{q}_j\right) = 1.
        \label{eq:fdjkaajf}
    \end{equation}
    The {contrapositive} of \cref{eq:fdjkaajf} is
    \begin{equation}
        \varGamma^{NS}\left(\boldsymbol{p}_i,\boldsymbol{q}_j\right) = 0 
        \rightarrow
        \delta_e > 2r^{local}.
        \label{eq:cdjaj}
    \end{equation}
    Let $r^{local} \ge \frac{\tau_e}{2}$. Then if $\varGamma^{NS}\left(\boldsymbol{p}_i,\boldsymbol{q}_j\right) = 0$,
    we have $\delta_e > 2r^{local} \ge \tau_e$ (\cref{eq:cdjaj}), 
    and so the correspondence $(\boldsymbol{p}_i, \boldsymbol{q}_j)$ is an outlier (\cref{vgfjkd}), thus:
    \begin{equation}
        \forall r^{local} \ge \frac{\tau_e}{2}, \left( \varGamma^{NS}\left(\boldsymbol{p}_i,\boldsymbol{q}_j\right) = 0 \right) 
        \rightarrow
        \left( 
            \varOmega \left( \boldsymbol{p}_i,\boldsymbol{q}_j\right) = 0 
        \right).
        \label{eq:fnsstau}
    \end{equation}
    \cref{eq:fnsstau} signifies that when the search radius $r^{local}$ is sufficiently large (greater than $\frac{\tau_e}{2}$), a correspondence can be categorized as outlier (inter-class mismatching)  if two keypoints fail to meet the LS-Consistency.
    \label{proof:hjfdja}
\end{proof}
It is worth noting that the above reasoning does not consider the sparsity and noise from semantic segmentation in real point clouds. However, thanks to the robustness of Local-SS, \cref{eq:fnsstau} remains robust in real scenarios. For more details, please refer to \cref{sec_snsr} and \cref{sec_raslg}.

\section{Categories for the Construction of Landmarks}
When constructing landmarks for outdoor scenarios, some categories with high saliency scores are selected. Note that for indoor scenarios, landmarks are simply voxel-downsampled point clouds including all categories. Here, we explain how we select the semantic categories to guarantee the reliability of the BMR-SS. 
We first define the Ring-wise Semantic Consistency (RWS-Consistency) in \cref{def:df2}.

\begin{figure}[t!]
	\centering
	\includegraphics[width=0.7\linewidth]{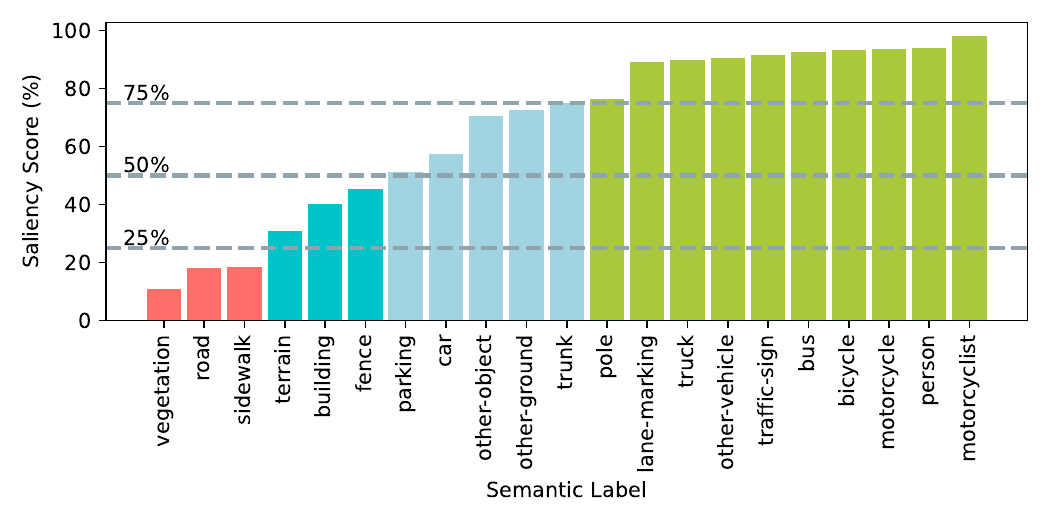}
	\caption{
		Saliency scores of different semantic labels when $r_{i-1}=10$m and $r_{i}=12$m in the Semantic-KITTI dataset.
	}
	\label{fig:03fsda}
\end{figure}

\begin{definition}[Ring-wise Semantic Consistency]
    Given a source keypoint $\boldsymbol{p}_i$ and a target keypoint $\boldsymbol{q}_j$.
    If both keypoints share the same label $s_t$ in the $k$-th ring, they are considered to satisfy RWS-consistency under semantic label $s_t$ in $k$-th ring, mathematically:
    \begin{equation}
		s_t\in \left( \varUpsilon _{k}^{\tilde{\mathcal{P}}}\left( \boldsymbol{p}_i \right) \cap \varUpsilon _{k}^{\tilde{\mathcal{Q}}}\left( \boldsymbol{q}_j \right) \right) \rightarrow \varGamma _{tk}^{SS}\left( \boldsymbol{p}_i,\boldsymbol{q}_j \right) =1, 
        \label{eq:E7IS23}
    \end{equation}
	where
	$\tilde{\mathcal{P}}$ and $\tilde{\mathcal{Q}}$ represent the source and target landmark sets, respectively.
	$\varUpsilon^{\tilde{\mathcal{P}}}_k(\boldsymbol{p}_i)$ and $\varUpsilon^{\tilde{\mathcal{Q}}}_k(\boldsymbol{q}_j)$ retrieve the label sets of source and target landmarks within the $k$-th ring around $\boldsymbol{p}_i$ and $\boldsymbol{q}_j$, respectively.
	$\varGamma_{kt}^{SS} : \mathcal{P} \times \mathcal{Q} \rightarrow \{0,1\}$ is an indicator function of whether $\boldsymbol{p}_i \in \mathcal{P}$ and $\boldsymbol{q}_j \in \mathcal{Q}$ satisfy the RWS-Consistency in the $k$-th ring under semantic label $s_t$.
    \label{def:df2}
\end{definition}
\cref{eq:E7IS23} provides clues of similar position (scene similarity) by checking the occurrence of $s_{t}$ in the $k$-th ring. However, the positioning performance of different categories of landmarks varies.
Generally, landmarks that appear less frequently in a scene are more effective as key points of reference due to their uniqueness.
For example, we usually choose to use bus stations instead of roads for localization. Hence, we define a saliency matrix $\mathbf{W}$ that measures the localization abilities of $s_t$ in the $k$-th ring:
\begin{equation}
	\left[ \textbf{W} \right] _{tk}=1-\frac{\left| \bigcup_{\tilde{\boldsymbol{p}}\in \tilde{\mathcal{P}}^{s_t}}{\varPsi _{k}^{\hat{\mathcal{P}}}\left( \tilde{\boldsymbol{p}} \right)} \right|}{\left| \hat{\mathcal{P}} \right|},
	\label{eq:4zm7U2}
\end{equation}
where $[\mathbf{W}]_{tk}$ denotes the element in the $t$-th row and $k$-th column of $\mathbf{W}$.
$
\tilde{\mathcal{P}}^{s_t}=\{ \tilde{\boldsymbol{p}}_i|\rho ^{\mathcal{P}}\left( \tilde{\boldsymbol{p}}_i \right) =s_t,\ \tilde{\boldsymbol{p}}_i\in \tilde{\mathcal{P}} \} 
$
is a landmark set of category $s_t$, where $\tilde{\mathcal{P}}$ is {a source landmark set}.
$\rho^{\mathcal{P}}(\cdot)$ is a label mapping function.
$
\varPsi _{k}^{\hat{\mathcal{P}}}\left( \tilde{\boldsymbol{p}} \right) 
$ retrieves subpoint set of raw source point cloud $\hat{\mathcal{P}}$ in the $k$-th ring of landmark $\tilde{\boldsymbol{p}_i}$.
The numerator includes the subpoint set in $\hat{\mathcal{P}}$ satisfying RWS-Consistency with $\boldsymbol{p}_i$.
The denominator represents the total number of points in $\hat{\mathcal{P}}$.
\cref{fig:03fsda} visualizes the saliency score of different types of $s_t$ in the ring of radii ranging from $10m$ to $20m$, studied based on the Semantic-KITTI dataset. It is evident that the saliency of frequently occurring categories (e.g., roads and vegetation) is relatively poor, while entities with high uniqueness (e.g., traffic lights) show better localization performance in the scene.

Based on the saliency score matrix, we select the categories for our RWS-Consistency. 
Specifically, we first remove categories with saliency less than $50\%$ and dynamic categories (such as bus, bicycle, motorcycle, person, motorcyclist). Please note that the car and truck categories are retained because most instances of these two categories are static, which is helpful for registration. Finally, the following five categories are selected: trunk, truck, pole, traffic-sign, and car.

\section{More Registration Comparasion}
\label{sec:high_reg}

\para{Highway Registration Comparison.}
The results for the highway sequence are displayed in Tab.~\ref{tab:tabhi} and~\ref{tab:tabhi2}.
Regardless of the descriptor used, our method significantly enhances the IN/IR across all datasets.
It is noteworthy that the improvement gap is much larger than when considering all sequences.
This is primarily due to the highway scenario's inherently low distinguishability of local geometric patches, leading to a drastic downgrade in local descriptor performance.
However, ML-SemReg relies solely on landmarks, which enhances robustness to local geometric similarities.

\begin{table*}[t!]
    \renewcommand{\arraystretch}{1.2}
    \centering
    \renewcommand\tabcolsep{2.5pt}
    \caption{Keypoint matching comparison on sequence~1 of the KITTI (highway).}
    \resizebox{\linewidth}{!}{
        \begin{tabular}{c|cc|cc|cc|cc|cc|cc}
            \hline
            \multirow{3}[1]{*}{Matcher} & \multicolumn{6}{c|}{FPFH}  & \multicolumn{6}{c}{GEDI}                                                                                                                                                                                                                                                \\ \cline{2-13}
                                        & \multicolumn{2}{c|}{easy}  & \multicolumn{2}{c|}{medium} & \multicolumn{2}{c|}{hard} & \multicolumn{2}{c|}{easy} & \multicolumn{2}{c|}{medium} & \multicolumn{2}{c}{hard}                                                                                                                            \\
            ~                           & IN                         & IR(\%)                      & IN                        & IR(\%)                    & IN                          & IR(\%)                   & IN                  & IR(\%)            & IN                 & IR(\%)            & IN                 & IR(\%)           \\ \hline
            MNN                         & 55.45                      & 4.56                        & 19.65                     & 1.76                      & 8.65                        & 0.80                     & 268.30              & 18.15             & 117.71             & 9.00              & 51.90              & 4.29             \\
            NN                          & \multicolumn{1}{c}{147.26} & 2.95                        & 58.48                     & 1.17                      & 28.65                       & 0.57                     & 584.39              & 11.69             & 279.18             & 5.58              & 129.05             & 2.58             \\
            OT                          & 141.14                     & 2.82                        & 55.18                     & 1.10                      & 28.40                       & 0.57                     & 580.24              & 11.60             & 273.45             & 5.47              & 128.05             & 2.56             \\
            \hline
            MNN+Ours                    & 255.33                     & \textbf{20.12}              & 110.33                    & \textbf{10.44}            & 44.65                       & \textbf{5.11}            & 562.34              & \textbf{37.51}    & 271.11             & \textbf{22.56}    & 119.11             & \textbf{11.29}   \\
            NN+Ours                     & \textbf{670.99}            & \underbar{11.00}            & \underline{322.99}        & \underline{5.35}          & \underline{154.44}          & {2.51}                   & \underline{1264.96} & \underline{21.77} & \underline{680.11} & 11.33             & \underline{321.79} & 5.33             \\
            OT+Ours                     & \underline{668.11}         & 10.85                       & \textbf{331.11}           & 5.34                      & \textbf{158.54}             & \underline{2.69}         & \textbf{1300.14}    & 22.00             & \textbf{701.21}    & \underline{11.41} & \textbf{333.38}    & \underline{5.41} \\
            \hline
        \end{tabular}
    }
    \label{tab:tabhi}
\end{table*}

\begin{table*}[th]
    \centering
    \renewcommand{\arraystretch}{1.2}
    \renewcommand\tabcolsep{2.5pt}
    \caption{Registration results on sequence~1 of the KITTI (highway).}
    \resizebox{\linewidth}{!}{
        \begin{tabular}{c|c|ccc|ccc|ccc}
            \hline
            \multirow{2}[1]{*}{Descriptor} & \multirow{2}[1]{*}{Estimator}       & \multicolumn{3}{c|}{easy} & \multicolumn{3}{c|}{medium} & \multicolumn{3}{c}{hard}                                                                                                                       \\
                                           &                                     & RE($^\circ$)              & TE(cm)                      & RR(\%)                   & RE($^\circ$)     & TE(cm)            & RR(\%)            & RE($^\circ$)     & TE(cm)            & RR(\%)            \\ \hline
            \multirow{12}[1]{*}{FPFH}      & RANSAC1M \cite{fischler1981random}  & 1.96                      & 52.37                       & 1.75                     & -                & -                 & 0.00              & -                & -                 & 0.00              \\
                                           & RANSAC4M \cite{fischler1981random}  & \multicolumn{1}{c}{1.01}  & 28.57                       & 7.46                     & 0.72             & 39.07             & 2.59              & -                & -                 & 0.00              \\
                                           & TeaserPlus  \cite{Yang20tro-teaser} & 0.54                      & 17.40                       & 28.07                    & 1.04             & 27.72             & 10.34             & 0.95             & 45.60             & 2.56              \\
                                           & PointDSC  \cite{bai2021pointdsc}    & 0.53                      & 18.14                       & 41.23                    & 1.06             & 33.10             & 11.21             & -                & -                 & 0.00              \\
                                           & MAC  \cite{zhang20233d}             & {0.22}                    & 9.81                        & 45.18                    & 0.54             & 19.17             & 23.28             & 0.47             & \textbf{17.49}    & 2.56              \\
                                           & SC2PCR  \cite{Chen_2022_CVPR}       & 0.25                      & 10.12                       & 42.54                    & 0.49             & 19.04             & 13.79             & 0.73             & 23.00             & 3.85              \\

            \cline{2-11}
                                           & RANSAC1M+Ours                       & 1.63                      & 35.00                       & 43.82                    & 1.72             & 36.77             & 15.55             & 1.41             & 32.60             & 5.17              \\
                                           & RANSAC4M+Ours                       & 1.48                      & 33.46                       & 49.90                    & 1.70             & 36.05             & 23.34             & 1.43             & 40.05             & 7.71              \\
                                           & TeaserPlus+Ours                     & 0.28                      & 10.29                       & \underline{73.68}        & 0.61             & 19.52             & 58.84             & 0.98             & 27.13             & 33.43             \\
                                           & PointDSC+Ours                       & 0.30                      & 8.38                        & \textbf{79.39}           & 0.49             & 14.55             & {62.96}           & 0.68             & 23.36             & 38.54             \\
                                           & MAC+Ours                            & \underline{0.17}          & \underline{6.57}            & \textbf{79.39}           & \textbf{0.30}    & \underline{12.02} & \underline{64.69} & \underline{0.45} & {19.34}           & \underline{49.83} \\
                                           & SC2PCR+Ours                         & \textbf{0.16}             & \textbf{6.37}               & \textbf{79.39}           & \underline{0.31} & \textbf{10.97}    & \textbf{64.84}    & \textbf{0.38}    & \underline{19.29} & \textbf{50.10}    \\
            \hline
            \hline
            \multirow{12}[1]{*}{FPFH}      & RANSAC1M \cite{fischler1981random}  & 0.91                      & 25.76                       & 41.23                    & 0.78             & 26.07             & 12.93             & 1.43             & 52.66             & 1.28              \\
                                           & RANSAC4M \cite{fischler1981random}  & 1.03                      & 27.93                       & 53.95                    & 1.10             & 27.33             & 22.41             & 2.66             & 28.01             & 2.56              \\
                                           & TeaserPlus  \cite{Yang20tro-teaser} & 0.20                      & 8.18                        & 66.67                    & 0.46             & 14.45             & 37.07             & 0.71             & 20.93             & 21.79             \\
                                           & PointDSC  \cite{bai2021pointdsc}    & 0.19                      & 7.47                        & 71.93                    & 0.40             & 14.79             & 59.48             & 0.62             & 26.11             & 24.36             \\
                                           & MAC  \cite{zhang20233d}             & \underline{0.14}          & 4.70                        & 72.81                    & 0.31             & 10.02             & 68.26             & 0.43             & 17.03             & 29.59             \\
                                           & SC2PCR  \cite{Chen_2022_CVPR}       & \underline{0.14}          & 4.73                        & 72.37                    & 0.25             & 8.53              & 60.34             & 0.37             & 15.59             & 33.33             \\

            \cline{2-11}
                                           & RANSAC1M+Ours                       & 1.12                      & 25.37                       & 68.54                    & 1.08             & 29.52             & 39.80             & 1.15             & 30.33             & 10.38             \\
                                           & RANSAC4M+Ours                       & 1.05                      & 28.05                       & 65.85                    & 1.38             & 32.74             & 51.32             & 1.41             & 37.02             & 20.64             \\
                                           & TeaserPlus+Ours                     & 0.19                      & 8.09                        & {86.34}                  & 0.34             & 12.73             & 75.54             & 0.52             & 22.59             & 57.70             \\
                                           & PointDSC+Ours                       & 0.21                      & 6.25                        & \textbf{88.66}           & \textbf{0.23}    & 10.86             & {85.58}           & 0.37             & 19.45             & \underline{63.37} \\
                                           & MAC+Ours                            & \textbf{0.12}             & \underline{4.42}            & \underline{87.78}        & \textbf{0.23}    & \textbf{7.63}     & \underline{86.00} & \underline{0.32} & \textbf{12.46}    & {62.06}           \\
                                           & SC2PCR+Ours                         & \textbf{0.12}             & \textbf{4.31}               & {86.45}                  & \underline{0.28} & \underline{7.72}  & \textbf{86.34}    & \textbf{0.31}    & \underline{12.58} & \textbf{70.97}    \\

            \hline
        \end{tabular}

    }
    \label{tab:tabhi2}
\end{table*}

\begin{table}[t!]
    \centering
    \renewcommand{\arraystretch}{1.2}
    \renewcommand\tabcolsep{2.5pt}
    \caption{Registration results compared to the state-of-the-art deep learning framework in sequences 8, 9, and 10.}
    \resizebox{\linewidth}{!}{
        \begin{tabular}{cc|ccc|ccc|ccc}
            \hline
            \multirow{2}[1]{*}{Descriptor}               & \multirow{2}[1]{*}{Registrator}                            & \multicolumn{3}{c|}{easy} & \multicolumn{3}{c|}{medium} & \multicolumn{3}{c}{hard}                                                                                                                       \\
                                     &                                        & RE(°)                     & TE(cm)                      & RR(\%)                   & RE(°)            & TE(cm)            & RR(\%)            & RE(°)            & TE(cm)            & RR(\%)            \\
            \hline
            -                        & GeoTransformer \cite{qin2022geometric} & 0.37                      & 9.10                        & \textbf{99.82}           & 0.68             & 24.62             & 15.66             & 0.77             & 28.12             & 1.08              \\
            -                        & CoFiNet \cite{yu2021cofinet}           & 0.41                      & 9.29                        & \textbf{99.82}           & 1.49             & 27.21             & 46.98             & 1.89             & 29.03             & 1.08              \\
            \hline
            \multirow{4}[1]{*}{FPFH} & SC2PCR                                 & 0.38                      & 11.00                       & 96.76                    & 0.92             & 24.33             & 68.39             & 1.27             & 25.98             & 7.84              \\
                                     & PointDSC                               & 0.38                      & 11.23                       & 95.44                    & 0.93             & 24.38             & 55.48             & 1.31             & 26.83             & 3.92              \\
                                     & SC2PCR+Ours                            & 0.25                      & 6.81                        & 99.10                    & 0.50             & 14.34             & 94.19             & 0.75             & 20.74             & 71.57             \\
                                     & PointDSC+Ours                          & 0.25                      & 6.71                        & 99.28                    & 0.48             & 14.27             & 93.55             & 0.69             & 19.12             & 70.59             \\
            \hline
            \multirow{4}[1]{*}{GEDI} & SC2PCR                                 & \textbf{0.23}             & \underline{5.19}            & \underline{99.64}        & 0.40             & 12.03             & 94.84             & 0.65             & 18.52             & 62.75             \\
                                     & PointDSC                               & \textbf{0.23}             & \textbf{5.17}               & \underline{99.64}        & \underline{0.39} & 12.15             & \underline{95.48} & \underline{0.58} & 19.22             & 50.00             \\
                                     & SC2PCR+Ours                            & \underline{0.24}          & 5.65                        & \underline{99.64}        & 0.40             & \underline{11.57} & \textbf{98.06}    & \textbf{0.57}    & \textbf{17.38}    & \textbf{85.29}    \\
                                     & PointDSC+Ours                          & \textbf{0.23}             & 5.58                        & \underline{99.64}        & \textbf{0.37}    & \textbf{11.49}    & \textbf{98.06}    & \textbf{0.57}    & \underline{18.82} & \underline{81.37} \\
            \hline
        \end{tabular}
    }
    \label{tab:chdja}
\end{table}

\para{Deep-learning Framework Comparasion.}
\label{sec:deep_com}
We also compare our algorithm against current state-of-the-art deep registration methods, i.e., GeoTransformer \cite{qin2022geometric} and CoFiNet \cite{yu2021cofinet}, in sequences 8, 9, and 10.
These sequences are commonly used for testing in deep registration frameworks \cite{qin2022geometric, yu2021cofinet}.
As indicated in \cref{tab:chdja}, almost all methods reach saturation in the easy case, yet our method attains higher registration precision. As the overlap ratio decreases, our method consistently demonstrates robustness even at low overlap ratios. In contrast, the deep learning framework deteriorates due to its limited generalization ability.

\section{Robustness Analysis for Semantic Label Generation}

\begin{figure}
	\centering
	\includegraphics[width=.5\linewidth]{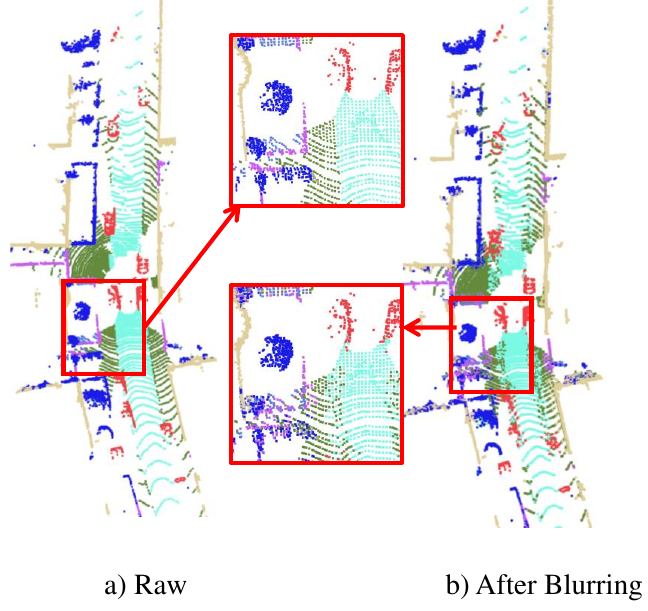}
	\caption{Visualization of semantic label degeneration with boundary ambiguity (BR=$100cm$).}
	\label{fig2fjkdajkfda}
\end{figure}

\label{sec_raslg}

Semantic boundary ambiguity \cite{gong2021boundary, tang2022contrastive} may affect the performance of our method, as semantic boundaries often exhibit distinguishable geometric features.
To evaluate the robustness of ML-SemReg to semantic boundary ambiguity, we degraded the semantic labels by blur boundary.
Specifically, we set the label of each boundary point with a probability of $0.5$, to a random other semantic class within its radius (denoted as Blur Radius, BR).
\cref{fig2fjkdajkfda} shows the case of BR=$100cm$.
The results of correspondences are shown in \cref{tab1}, we can conclude that GM/MM and ML-SemReg (GM+MM) exhibit robustness to degradation of semantic,
e.g., when BR=$100 cm$, the IR of ML-SemReg drops by less than $2\%$ compared to using raw semantic labels.
\begin{table}[t]
	\centering
	\renewcommand{\arraystretch}{1}
	\renewcommand\tabcolsep{2pt}
	\caption{The correspondence results under semantic label degradation. `-' represents not blurring semantic labels.}
	\resizebox{\linewidth}{!}{
		\begin{tabular}{c|cc|cc|cc|cc|cc|cc}
			\hline
			\multicolumn{1}{c|}{BR} & \multicolumn{2}{c|}{-} & \multicolumn{2}{c|}{40 cm} & \multicolumn{2}{c|}{60 cm} & \multicolumn{2}{c|}{80 cm} & \multicolumn{2}{c|}{100 cm} & \multicolumn{2}{c}{120 cm}                                                          \\
			\hline
			Metrics                 & IN                     & IR(\%)                     & IN                         & IR(\%)                     & IN                          & IR(\%)                     & IN      & IR(\%) & IN      & IR(\%) & IN      & IR(\%) \\
			\hline
			Group Matching                     & 581.30                 & 9.67                       & 583.69                     & 9.51                       & 589.33                      & 9.37                       & 595.05  & 9.14   & 596.27  & 8.99   & 605.91  & 8.77   \\
			MM                     & 821.77                 & 16.39                      & 802.43                     & 16.05                      & 793.84                      & 15.95                      & 777.13  & 15.51  & 746.49  & 15.01  & 706.11  & 14.14  \\
			ML-SemReg               & 1146.90                & 18.93                      & 1154.90                    & 18.68                      & 1169.31                     & 18.23                      & 1153.55 & 17.60  & 1149.08 & 16.95  & 1138.03 & 16.27  \\
			\hline
		\end{tabular}
	}
	\label{tab1}
\end{table}
\section{Selection of Neighborhood Search Radius}
\label{sec_snsr}

\begin{figure}[t!]
	\centering
	\includegraphics[width=.6\linewidth]{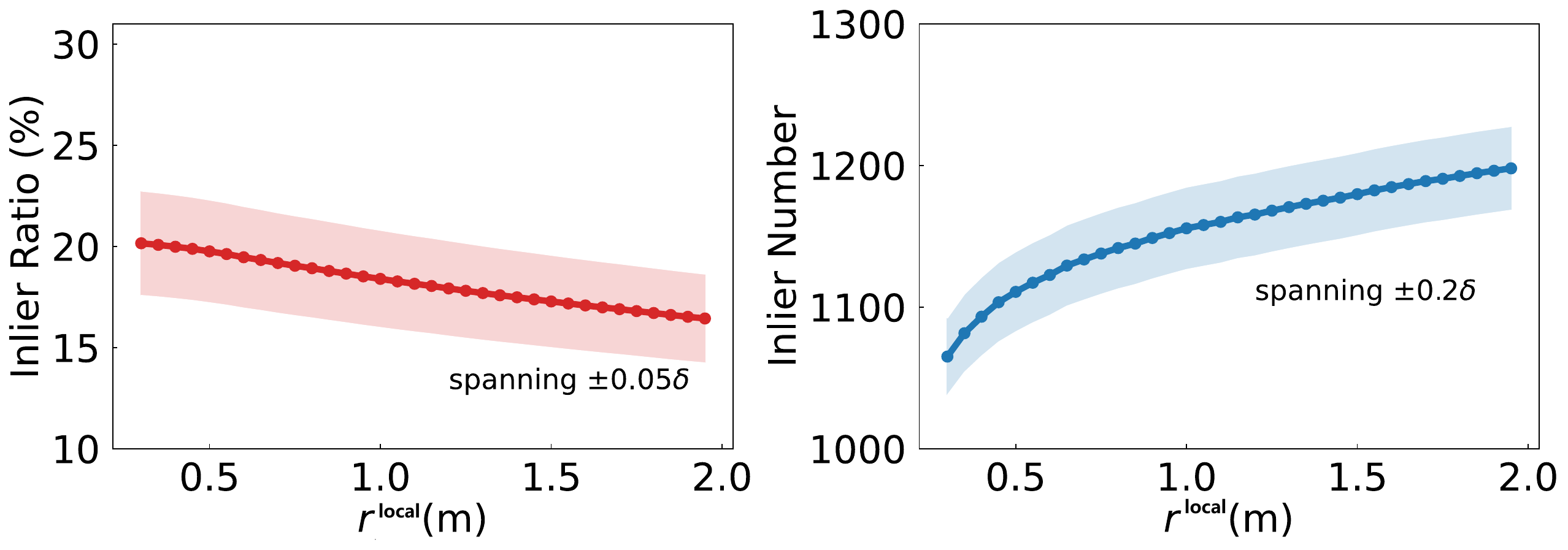}
	\caption{
		{
			Inlier ratio and number under different neighborhood search radius $r^{local}$.
		}
	}
	\label{fig1}
\end{figure}

We conducted an ablation study on $r^{local}$.
The results in \cref{fig1} revealed that as $r^{local}$ gradually increases from $0.25 m$ to $2.0 m$, the IR linearly decreases from $20.15\%$ to $16.43\%$, while the IN steadily rises from $1065$ to $1197$. This trend is attributed to the increased probability of inliers satisfying the LS-Consistency as $r^{local}$ increases, consequently encompassing more inliers.
However, this will also result in additional outliers being included, leading to a slight decline in the IR.

\section{Time cost analysis}

\begin{table}[th!]
	\centering
	\renewcommand{\arraystretch}{1.2}
	\renewcommand\tabcolsep{3.0pt}
	\caption{
	Average consumed time (ms) per point cloud pair on the KITTI dataset.
		{ML-SemReg}
	involves the construction of Local-SS/BMR-SS and matching using GM+MM (with NN) modules, thus
	\textit{Total-1=SUM(Local-SS, BMR-SS, GM+MM)}.
	{JUST-MM} matches keypoints using MM (with NN) with BMR-SS signature, which means \textit{Total-2=SUM(Local-SS, MM)}.
	}

	\resizebox{0.75\linewidth}{!}{
		\begin{tabular}{c|cc|cc|cc}
			\hline
			\multirow{2}{*}{Keypoints Number} & \multicolumn{2}{c|}{Signature} & \multicolumn{2}{c|}{ML-SemReg} & \multicolumn{2}{c}{JUST-MM}                              \\  
			\cline{2-7}
			                           & Local-SS                           & BMR-SS                        & GM+MM                      & Total-1 & MM    & Total-2 \\
			\hline
			250                        & 9.88                          & 6.67                          & 14.59                        & 31.14   & 4.02   & 10.69   \\
			500                        & 12.61                         & 5.98                          & 27.28                        & 45.87   & 14.03  & 20.02   \\
			1000                       & 18.15                         & 6.32                          & 58.93                        & 83.39   & 39.71  & 46.03   \\
			2500                       & 35.19                         & 6.72                          & 187.23                       & 229.14  & 204.04 & 210.76  \\
			5000                       & 62.87                         & 6.38                          & 513.56                       & 582.81  & 403.27 & 409.65  \\
			\hline
		\end{tabular}
	}
	\label{tab:sdcm}
\end{table}

We measure the runtime of ML-SemReg with different configurations on the KITTI dataset. 
As the MM module can achieve performance close to ML-SemReg, we also report results using only MM (denoted as {JUST-MM}).
The results are presented in \cref{tab:sdcm}, 
from which the following observations can be made:
(1) The time of ML-SemReg ranges from tens to hundreds of milliseconds, 
e.g., it only costs about $0.2$ seconds to process $2500$ correspondences.
(2) The runtime of JUST-MM is lower than ML-SemReg. For example, with an input containing $5000$ keypoints, the runtime decreases by $29.71\%$ compared to ML-SemReg. Additionally, for $500$ and $1000$ keypoints per point cloud, MM runs at $50.0$ and $21.7$ 
FPS, respectively.
(3) As keypoints increase from 250 to 5000, Local-SS construction time grows linearly, while BMR-SS construction time remains nearly unchanged, at approximately 6.4 milliseconds.
\section{Visualization Results}
\label{sec:app_vis}
\begin{figure*}[t!]
	\centering
	\includegraphics[page=1,width=0.999999\linewidth]{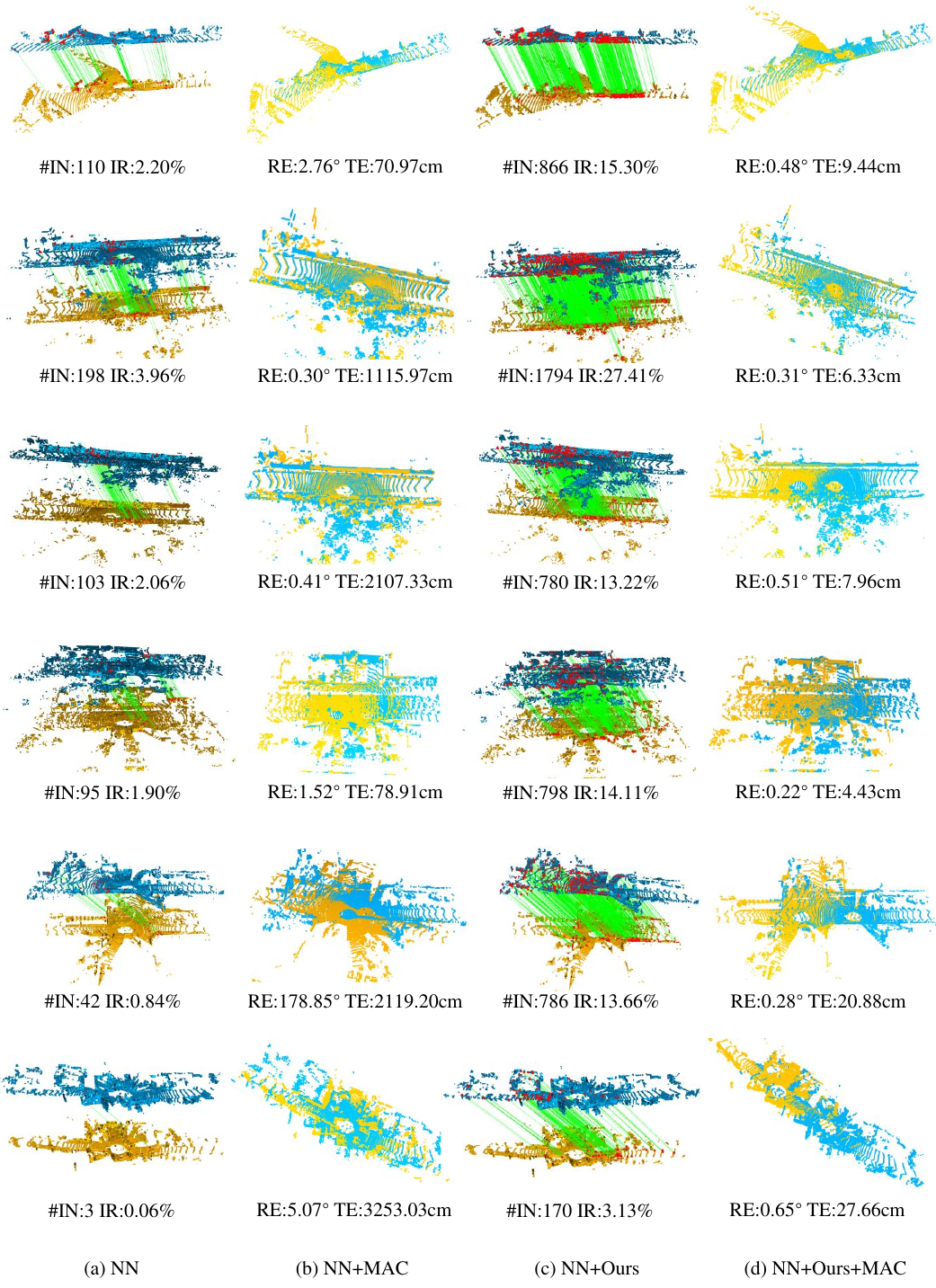}
	\caption{Qualitative comparison on the KITTI dataset. Rows 1 and 2 correspond to the easy dataset, Rows 3 and 4 to the medium dataset, and Rows 5 and 6 to the hard dataset.}
	\label{fig:supp_corres}
\end{figure*}

\cref{fig:supp_corres} visualizes correspondences and registration results obtained using NN and NN+Ours with the MAC estimator and FPFH descriptor.
Evidently, our method generates more inliers, which provides a denser and more dispersive set of inliers, significantly enhancing registration recall and precision.

\begin{figure*}[t!]
	\centering
	\includegraphics[page=2,width=0.999999\linewidth]{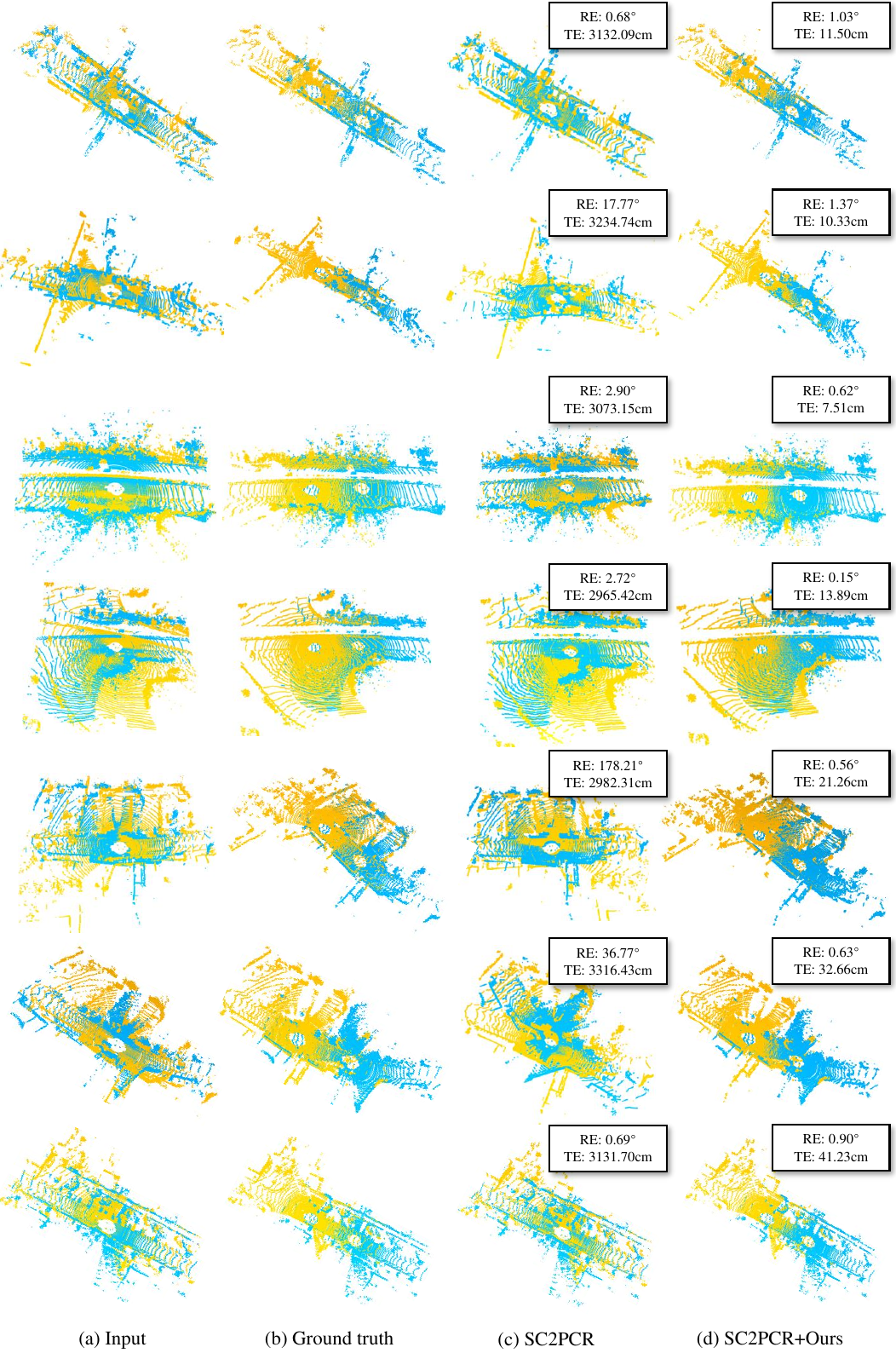}
	\caption{Registration results on the KITTI-hard (30-40m) dataset.}
	\label{fig_supp2}
\end{figure*}

\begin{figure*}[t!]
	\centering
	\includegraphics[page=1,width=0.99999\linewidth]{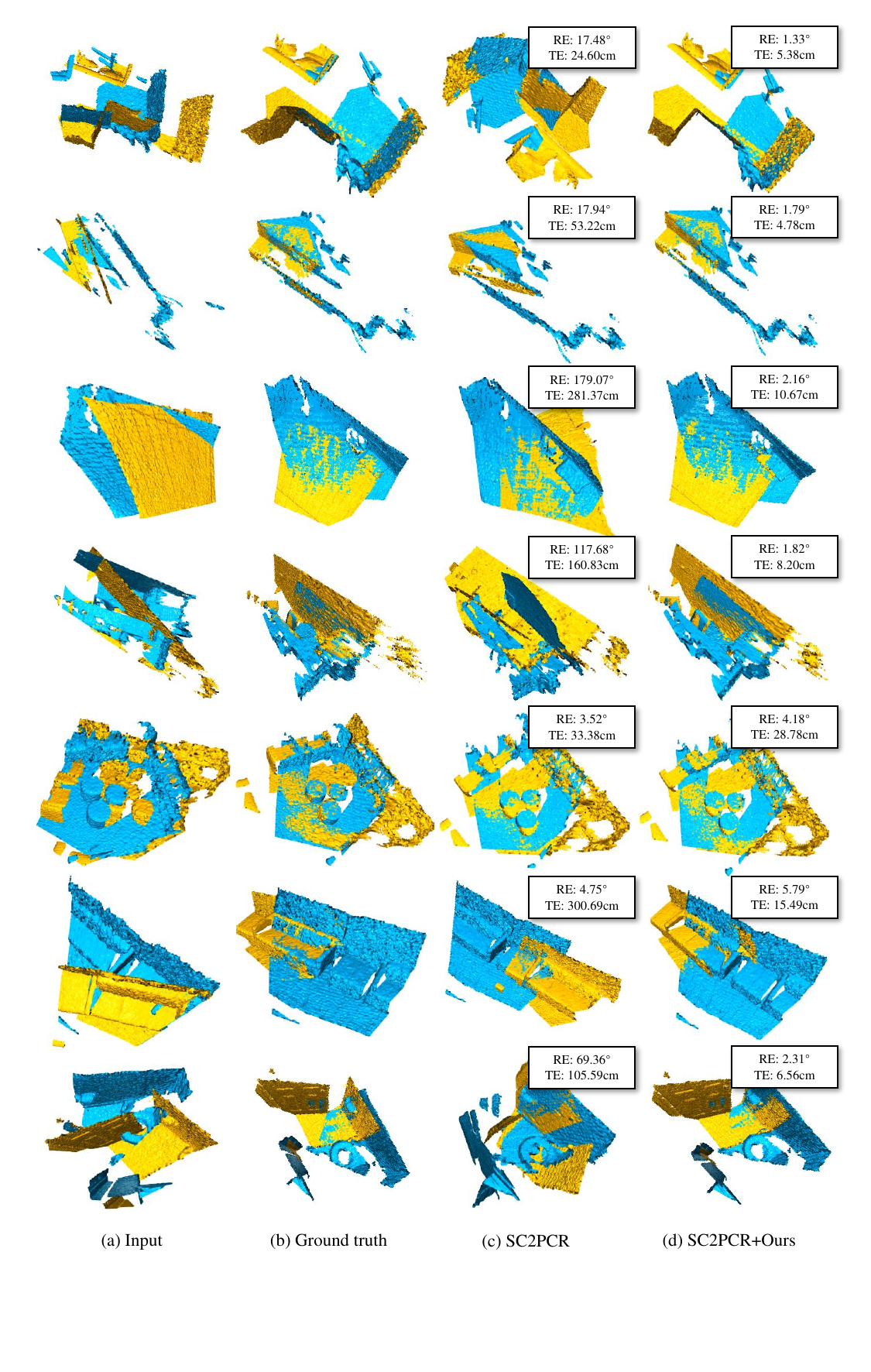}
	\caption{Registration results on the ScanNet (40\% $\sim$ 50\%) dataset.}
	\label{fig:supp1}
\end{figure*}

\cref{fig:supp1,fig_supp2}  illustrate the registration results of SC2PCR and SC2PCR+Ours on KITTI-hard (30-40m) and ScanNet ($40\%\sim 50\%$). It is evident that SC2PCR exhibits significant improvement with the utilization of high-quality correspondences generated by NN+Ours.

\end{sloppypar} 
\end{document}